\documentclass[letterpaper]{article} 
\usepackage{aaai24}  
\usepackage{times}  
\usepackage{helvet}  
\usepackage{courier}  
\usepackage[hyphens]{url}  
\usepackage{graphicx} 
\usepackage{natbib}  
\usepackage{caption} 
\usepackage{algorithmic}
\usepackage{amssymb}
\usepackage{amstext}
\usepackage{amsmath}
\usepackage{booktabs}
\usepackage[ruled,vlined]{algorithm2e}
\usepackage{xcolor}

\usepackage{multirow}
\usepackage{graphicx}
\usepackage{subcaption}

\title{Transformer-based No-Reference Image Quality Assessment via Supervised Contrastive Learning}
\author{
    Jinsong Shi,
    Pan Gao\thanks{Corresponding author},
    Jie Qin
}
\affiliations{
    College of Computer Science and Technology, Nanjing University of Aeronautics and Astronautics\\

    \{srache, pan.gao\}@nuaa.edu.cn, qinjiebuaa@gmail.com
}

\usepackage{bibentry}

\begin{document}

\maketitle

\begin{abstract}
Image Quality Assessment (IQA) has long been a research hotspot in the field of image processing, especially No-Reference Image Quality Assessment (NR-IQA). Due to the powerful feature extraction ability, existing Convolution Neural Network (CNN) and Transformers based NR-IQA methods have achieved considerable progress. However, they still exhibit limited capability when facing unknown authentic distortion datasets. To further improve NR-IQA performance, in this paper, a novel supervised contrastive learning (SCL) and Transformer-based NR-IQA model SaTQA is proposed. We first train a model on a large-scale synthetic dataset by SCL (no image subjective score is required) to extract degradation features of images with various distortion types and levels. To further extract distortion information from images, we propose a backbone network incorporating the Multi-Stream Block (MSB) by combining the CNN inductive bias and Transformer long-term dependence modeling capability. Finally, we propose the Patch Attention Block (PAB) to obtain the final distorted image quality score by fusing the degradation features learned from contrastive learning with the perceptual distortion information extracted by the backbone network. Experimental results on seven standard IQA datasets show that SaTQA outperforms the state-of-the-art methods for both synthetic and authentic datasets. 
Code is available at \url{https://github.com/I2-Multimedia-Lab/SaTQA}
\end{abstract}

\section{Introduction}
 Image Quality Assessment (IQA) refers to the quantitative analysis of the content of an image, thus quantifying the degree of visual distortion of a distorted image. The relevant evaluation methods are generally divided into two types: subjective quality assessment~\cite{mantiuk2012comparison} and objective quality assessment~\cite{yang2009objective}. Among them, the objective quality assessment is divided into three categories according to the extent of information the algorithm needs from reference image ~\cite{wang2006modern}: full reference quality assessment (FR-IQA)~\cite{wang2004image}, reduced-reference quality assessment (RR-IQA)~\cite{rehman2012reduced} and no reference quality assessment (NR-IQA)~\cite{mittal2012no}. The FR-IQA and RR-IQA methods calculate the image quality by comparing the difference between the reference image and the distorted image, while the NR-IQA method evaluates the image quality based on the distorted image itself without any reference information, making it the most challenging task and more promising in practical applications. 

\begin{figure}[]
	\centering
	\includegraphics[scale=0.31]{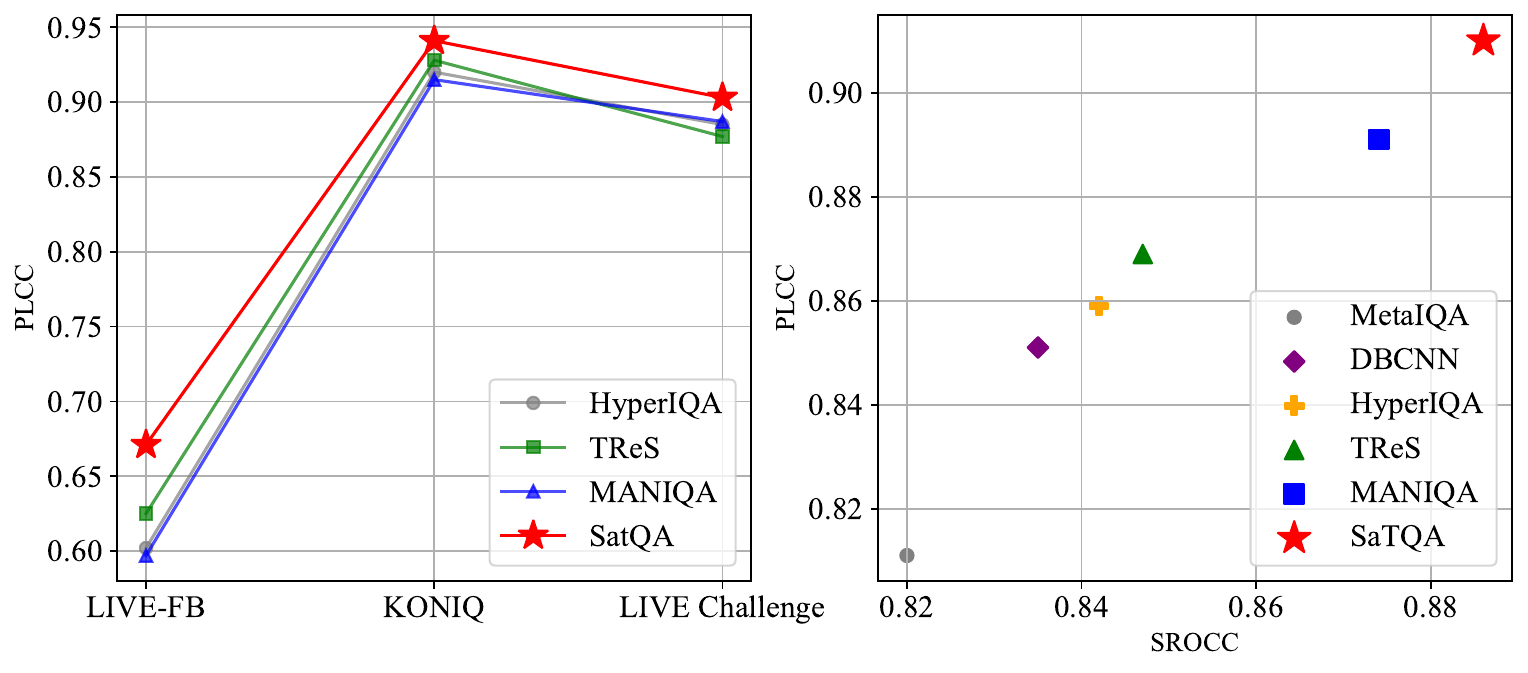}
	\caption{Quantitative comparison of NR-IQA methods. The left figure presents our PLCC results on three authentic datasets compared to the latest NR-IQA method. The right figure presents the average performance of our method compared to other methods on seven datasets. Higher coefficient matches perceptual score better.}
	\label{fig:fig0}
    \vspace{-2mm}
\end{figure}

\begin{figure}[]
	\centering
	\includegraphics[scale=0.65]{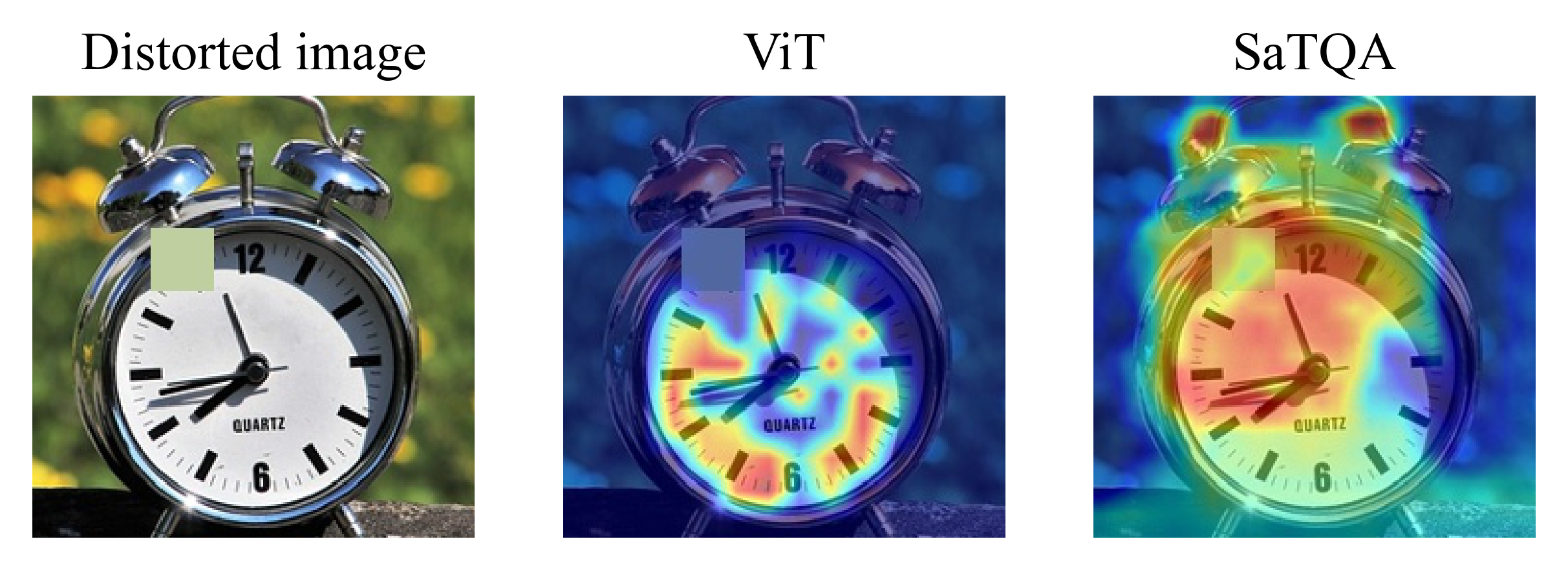}
	\caption{Grad-CAM~\cite{selvaraju2017grad} of ViT and SaTQA on distorted image.}
	\label{fig:cam_compare}
\end{figure}

With recent advance of deep learning techniques in computer vision, more and more data-driven approaches have been developed for NR-IQA \cite{mittal2012no, zhang2015feature, xue2013learning,golestaneh2022no,yang2022maniqa}. However, most existing deep learning based NR-IQA models are trained using the whole training dataset without distinguishing different distortion types and levels, which may render them exhibiting limited capability when facing a large-scale authentic dataset, such as LIVE-FB \cite{ying2020patches}. 
As shown in Fig\;\ref{fig:fig0}, even the recent most powerful Transformer-based NR-IQA model MANIQA \cite{yang2022maniqa}, the PLCC result on LIVE-FB is down to less than 0.6. 
This is because, for human eyes, even a same level of distortion but induced by two different types of noises may result in very different visual perception. Thus, without distinguishing different noise types and levels during training, the model may not perform well in the face of a large-scale dataset containing excessive distortions. Some CNN-based methods, such as CONTRIQUE \cite{madhusudana2022image},  attempted to solve this problem by specifying the distortion type and level or using other type of content as the label for distortion classification during training. However, this strategy is not applicable to large-scale authentic dataset, where the distortion type and level are impossible to explicitly specify. 
To tackle this problem, we propose a supervised contrastive learning based quality assessment method. By conducting self-supervised pre-training on the large scale synthetic dataset KADIS~\cite{lin2001weak}, degradation features corresponding to different types and levels of distorted images are obtained as prior knowledge for final quality score regression.
In the architecture design of IQA model, traditional methods usually employ CNN for perceptual distortion feature extraction~\cite{saad2012blind, mittal2012no, zhang2015feature, xue2013learning}. However, the perception of image quality by humans is usually performed by viewing the entire image, {i.e.}, by integrating the content distortion globally over the whole image. Due to limited receptive field of CNN, the performance of CNN based IQA models may be restricted. On the contrary to CNN, Transformer can integrate global distortion information of images due to its long range dependency modeling ability, which is thus naturally suitable for NR-IQA. Inspired by this, a plethora of Transformer based NR-IQA approaches have been proposed \cite{you2021transformer,ke2021musiq,golestaneh2022no,yang2022maniqa}. Nevertheless, pure Transformer has weak ability to extract local details such as image edges and textures, which may degrade the final prediction performance. In addition, most previous NR-IQA methods, such as MANIQA~\cite{yang2022maniqa}, used the pretrained Transformer as the backbone that is originally designed for image classification, which may not be well suitable for IQA that is closely related to human eye perception. 
As can be seen from the Grad-CAM~\cite{selvaraju2017grad} in Fig.\;\ref{fig:cam_compare}, pure ViT ~\cite{dosovitskiy2020image} model focuses more on the semantic region of the image, while missing the distorted areas of the image. To this end, we propose a Multi-Stream Block (MSB), which leverages the CNN and the Transformer features, aiming to combine the strengths of CNN's edge extraction while maintaining the global modeling capability of the Transformer. We further propose the Patch Attention Block (PAB) to fuse the learned perceptual feature with the degradation features from contrastive learning, and the fused features will be used for the prediction of objective image quality scores.

 The main contributions of this paper are as follows:

\begin{itemize}
	\item 
    We use supervised contrastive learning to perform self-supervised pre-training on the large-scale dataset KADIS, and the learned degradation features corresponding to various distortion types and levels of images are used to guide the training of the quality score generation network.
    \item
    We propose the MSB module that combines the features of CNN and Transformer, which introduces the inductive bias in CNN while ensuring the long-term dependency modeling capability of Transformer for enhancing the ability of the backbone network to extract perceptual distortion information.
	\item
    Extensive experiments on seven IQA datasets containing synthetic and authentic show that our proposed model outperforms current mainstream NR-IQA methods by a large margin.
\end{itemize}

\section{Related Work}

\begin{figure*}
	\centering
    \vspace{-2mm}
	\includegraphics[scale=0.8]{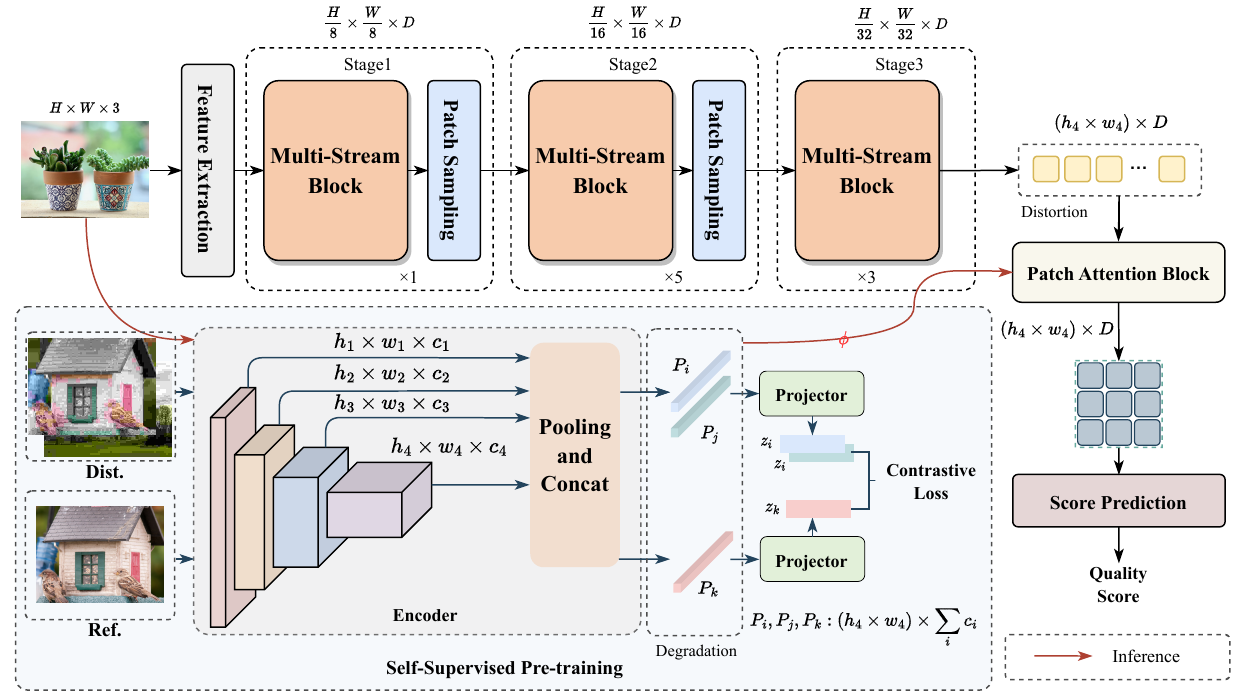}
	\caption{The overall framework of our approach for no-reference image quality assessment. $P_i$, $P_j$, and $P_k$ refer to the degradation features, $z_i$, $z_j$, and $z_k$ correspond to the vectors after projected. $\phi$ denotes stopping gradient propagation.}
	\label{fig:fig1}
	 \vspace{-4mm}
\end{figure*}

\subsection{No-Reference Image Quality Assessment}
The previous NR-IQA methods were mainly oriented to quality assessment task for specific distortion types~\cite{li2013referenceless, li2015no, liu2009no, wang2019blind}. These methods use statistical information on images to perform quality assessment of images with known distortion types. However, they are less effective when quality assessment is performed on images with unknown distortion types and therefore restrictive. Current NR-IQA methods focus on general-purpose, and can be further divided into two types according to the feature extraction method: natural scene statistics (NSS)-based metrics~\cite{gao2013universal, mittal2012no, moorthy2010two, saad2012blind, fang2014no} and learning-based metrics~\cite{ye2012no, ye2012unsupervised, zhang2015som, zhang2018blind, lin2018hallucinated, su2020blindly, zhu2020metaiqa}. The NSS-based methods consider that natural images have regular statistical features ({e.g.}, brightness, gradient, {etc.}), and images with different distortion types and levels will have different effects on this regularity. Based on this regulation, \cite{mittal2012no} used the NSS features in the spatial domain to construct the NR-IQA model. \cite{fang2014no} used the distribution function obtained by fitting the distorted image features to construct the evaluation model based on the moment characteristics of the image and the information entropy. \cite{moorthy2010two} used the discrete wavelet transform (DWT) to extract NSS features, while \cite{saad2012blind} used the discrete cosine transform (DCT) to evaluate the image quality. However, the NR-IQA methods based on NSS require manual extraction of local or global features of the image, and the feature representation extracted by this hand-crafted approach is often not effective. Although these NR-IQA methods show some performance gain on some synthetic distorted datasets, the results are not as good as on authentic distorted datasets. With the success of deep learning on various vision tasks, CNN-based NR-IQA methods have also achieved some success on distorted datasets. DBCNN~\cite{zhang2018blind} uses a CNN structure based on bilinear pooling for NR-IQA model building and achieves good performance gain on synthetic datasets. Hallucinated-IQA~\cite{lin2018hallucinated} proposes an NR-IQA model based on generative adversarial networks, which obtains the final quality score by pairing the generated hallucinated reference image with the distorted image. HyperIQA~\cite{su2020blindly} predicts image quality by fusing distortion information from images of different scales. MetaIQA~\cite{zhu2020metaiqa} trains the model to acquire prior knowledge of distorted images via meta-learning and solves the generalization performance problem of NR-IQA task to some extent.

\subsection{Transformer for IQA}
In the past few years, CNNs had become the popular backbone of computer vision tasks and achieved success in the field of IQA. Unfortunately, CNNs are highly flawed in capturing non-local information and have a strong localization bias. In addition, the shared convolutional kernel weights and translational invariance properties of CNNs also lead to the inability of CNNs to handle complex feature combinations. Inspired by the use of Transformer for long-range dependency modeling in the NLP domain~\cite{vaswani2017attention}, ViT~\cite{dosovitskiy2020image}, a Transformer model for computer vision tasks, has achieved initial success. In the field of IQA, IQT~\cite{you2021transformer} applies a hybrid architecture to extract feature information of reference and distorted images by CNN and use it as input to Transformer. MUSIQ~\cite{ke2021musiq} proposes a multi-scale image quality Transformer, which solves the problem of different image resolutions by encoding three scales of an image. TReS~\cite{golestaneh2022no} proposes to use relative ranking and self-consistency loss to train the model and thus enhance the evaluation performance. MANIQA~\cite{yang2022maniqa} proposes a multi-dimension attention mechanism for multi-dimensional interaction on channel and spatial domains, and achieves promising evaluation performance on synthetic datasets. However, all these Transformer based models are trained without distinguishing the latent degradation from other ones, which may not generalize well to a large-scale authentic dataset. 
\vspace{-1mm}

\subsection{Contrastive Learning}
Unlike supervised learning, self-supervised learning aims to acquire features using unlabeled data. 
Current contrastive learning methods mainly focus on instance recognition tasks, {i.e.}, the image itself and its corresponding enhanced samples are considered as the same class~\cite{dosovitskiy2014discriminative, chen2020simple, he2020momentum}. In addition, some self-supervised methods learn data features by constructing auxiliary tasks, which are often large and do not require manual annotation. Usually, these auxiliary tasks include rotation prediction~\cite{gidaris2018unsupervised}, image restoration~\cite{pathak2016context}, {etc.} \cite{liu2019exploiting} used image ranking as an auxiliary task and the proposed NR-IQA model achieved good performance on synthetic datasets. CONTRIQUE~\cite{madhusudana2022image} used distortion type and level as an auxiliary task by comparing user-generated content (UGC) with the images in synthetic datasets, finally, the learned features were used for image quality prediction. In this paper, we proposed a supervised contrastive learning (SCL) approach inspired by this auxiliary task construction method. But, taking one step further, we remove the use of UGC images and use the reference images from the synthetic dataset alone for comparasion.

\section{Proposed Method}
In this section, we detail our proposed model, which is an NR-IQA method based on supervised contrastive learning and Transformer named \textbf{SaTQA}, and Fig.\;\ref{fig:fig1} shows the architecture of our proposed model.

\subsection{Overall Pipeline}
Given a distorted image $I\in \mathbb{R}^{H\times W\times 3}$, where $H$ and $W$ denote height and width, respectively, and the goal of NR-IQA is to predict its objective quality score. First, we use ViT~\cite{dosovitskiy2020image} to extract the original features of the image, where each layer of extracted features defines $F_{i}\in \mathbb{R}^{(h\times w)\times C_i}$, where $i\in \left \{ 1, 2,...,12 \right \} $. In this paper, we extract four of these layers and concatenate them together by channel dimension, denoted as $\widehat{F}\in \mathbb{R}^{\sum_{i}{(h\times w)\times C_i}}$, where $i\in \left \{ 1,2,5,6 \right \} $, $C_i$ denotes the dimension of the ViT layer, $h$ and $w$ denotes the height and width of the corresponding output, respectively. Then we downscale $\widehat{F}$ along channel dimension to get $\widehat{F}^{'} \in \mathbb{R}^{h\times w\times D}$. Next we use a three-stage MSB module for image distortion extraction, where the first two stages are down-sampled after MSB processing. The distortion information is extracted by the backbone network and the output is $R\in \mathbb{R}^{(h_4\times w_4)\times D}$, where $D$ denotes the channel dimension of the features, $h_4$ and $w_4$ denotes the height and width of the last stage output. At the same time, the distorted image is passed through ResNet-50~\cite{he2016deep} trained with SCL and the output is $P$, $P\in \mathbb{R}^{(h_4\times w_4)\times \sum_{i} c_i}$. We make $P$ and $R$ have the same shape, and thus $D=\sum_{i} c_i$. Next, the PAB module fuses the features of $P$ and $R$ to obtain the feature $S$ for the image quality, and the final $S$ is regressed to the objective quality score corresponding to the image.

\subsection{Supervised Contrastive Learning\label{sec:3.2}}
Self-supervised learning improves model performance by augmenting the model's ability to extract data features through auxiliary tasks. In the NR-IQA domain, we need the model to distinguish the distortion types and levels of different images, and the target of this auxiliary task is thus similar to the classification of image. However, different from traditional supervised classification, we train the model using SCL on a large-scale synthetic dataset to obtain a more robust and generalized degradation features than classification, so that the degradation from the same distortion type (level) are pulled closer together than degradation from different distortion type (level). We then generalize the pre-trained model to other dataset with latent degradation, {e.g.}, authentic distortion dataset. As shown in Fig.\;\ref{fig:fig2}, at the beginning, the distribution of various distorted images on the feature space is random and unordered. Our goal is that images with the same distortion and level can be clustered together after using supervised contrastive learning.






\begin{figure}[]
	\centering
	\includegraphics[scale=0.37]{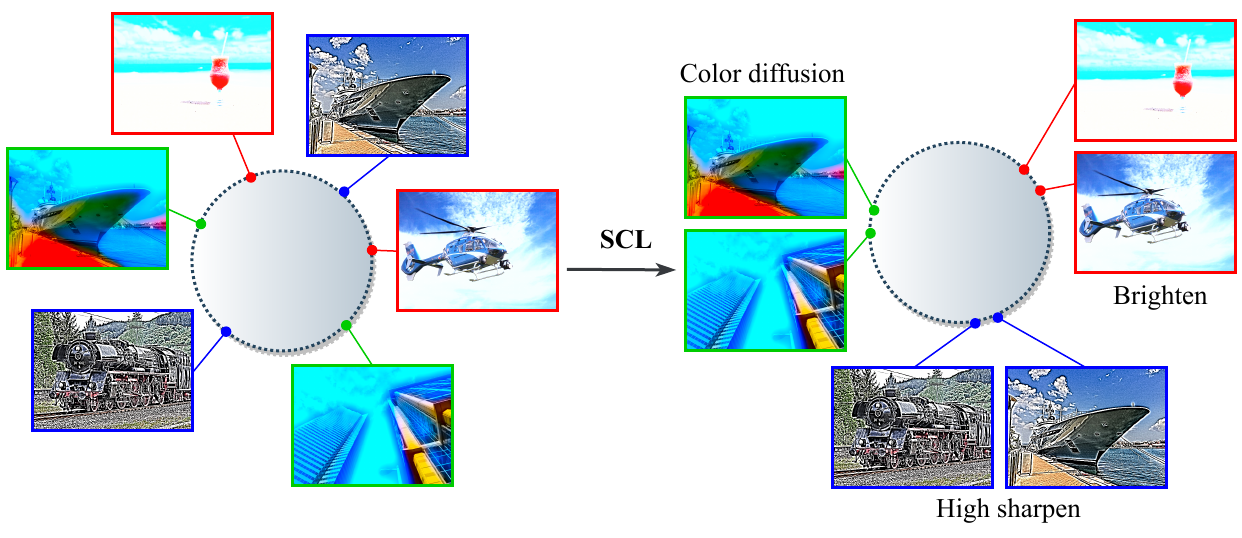}
	\caption{Left: Distribution of images with the same level of different distortion types in the feature space. Right: Distribution of images after training with SCL, where the images with color diffusion, high sharpen, brighten are clustered together, respectively.} 
	\label{fig:fig2}
\end{figure}

In our proposed supervised contrastive learning for distorted image clustering, firstly, $I_{u}^{v}$ is defined to denote the distorted image, $u\in \left \{ 1,...,U \right \} $ to denote the different distortion types corresponding to $I$, and $v\in \left \{ 1,...,V \right \} $ to denote the distortion level under each $u$, where $U$ and $V$ denote the total number of distortion types and levels. Thus, together with the class of the reference image, there are a total of $U\times V+1$ classes. That is, this auxiliary task is transformed into distinguish different types and levels of distorted images problem. Inspired by~\cite{chen2020simple}, we use encoder network $f(.)$ together with projector $g(.)$ structure to learn the degradation features of the images. In this paper, the encoder uses ResNet-50, and the projector uses a multilayer perceptron (MLP). For a given image $x$, $r=f(x)$, $z=g(r)$($z\in\mathbb{R}^{D}$), where $z$ denotes the $D$-dimensional feature of the output, $r$ denotes the feature representation of the encoder output. As with~\cite{chen2020simple}, the encoder output is $L_{2}$ normalized before being passed to the projector. The contrast loss function uses a supervised normalized temperature-scaled cross entropy (NT-Xent)~\cite{khosla2020supervised}, which can be formulated as: 
\begin{equation}
	\label{eq1}
	\mathcal{L}_{fea}=\frac{1}{|P(i)|} \sum_{j \in P(i)}-\log \frac{\exp (\phi(z_i, z_j) / \tau)}{\sum_{k=1}^N \mathbb{I}_{k \neq i} \exp (\phi(z_i, z_k) / \tau)}
\end{equation}
where $N$ denotes the number of all distortions and references images in a mini-batch, $\mathbb{I}$ denotes the indicator function, $\tau$ denotes the temperature parameter, $P(i)$ denotes the set belonging to the same class $i$, and $|P(i)|$ is its cardinality. $\phi(m, n)$ denotes $\phi(m, n)=m^{T} n /\|m\|_{2}\|n\|_{2}$.

\subsection{Multi-Stream Block\label{sec:3.3}}
As shown in Fig.\;\ref{fig:fig3} (a), we propose a Multi-Stream Block (MSB) that combines CNN to extract low-level edge features and Transformer to extract high-level content features of images. First, we reshape the feature $\widehat{F}'$ after ViT extraction into feature map $ \widetilde F \in \mathbb{R}^{D\times h\times w}$; then we divide it into three parts, $\chi_1\in \mathbb{R}^{D_1\times h\times w}$, $\chi_2\in \mathbb{R}^{D_2\times h\times w}$, and $\chi_3\in \mathbb{R}^{D_3\times h\times w}$, along the channel dimension, where $\chi_1$ and $\chi_2$ are feed to CNN and $\chi_3$ is feed to Transformer. $\chi_1$ will go through deformable convolution (DeformConv) and linear layers to get $\chi_1'$. $\chi_2$ will first go through depthwise convolution (DwConv) and max-pooling layers, and then up-sample to get$\chi_2'$. $\chi_3$ will go through Multi-Head Self-Attention (MHSA) and linear layers to get $\chi_3'$. These can be formulated as: 
\begin{align}
	& \chi_1'={\rm FC(Deform(Conv(}\chi_1)) \\
	& \chi_2'={\rm UpSample(MaxPool(DwConv(}\chi_2))) \\
	& \chi_3'={\rm FC(MHSA(}\chi_3))
\end{align}
Finally, $\chi_1'$, $\chi_2'$ and $\chi_3'$ are concatenated together and reshaped, which is denoted as $\widetilde{F}' \in \mathbb{R}^{D\times h\times w}$. In addition, we connect $\widetilde{F}$ to $\widetilde{F}'$ with residual learning through the CBAM~\cite{woo2018cbam} (\textit{i.e.}, channel attention and spatial attention) module. In this paper, the percentage of $\chi_1$, $\chi_2$ and $\chi_3$ in the channel dimension will change with different stages. At the beginning, all three share the same percentage in the channel, and then $\chi_3$ will gradually increase its percentage in the total channels. Note that, in CNN branches, deformable convolution can augment the spatial sampling locations, and depthwise convolution can reduce the number of computations and supplement the channel attention that is not modeled in Transformer branch. 
\begin{figure}[]
	\centering
	\includegraphics[scale=0.4]{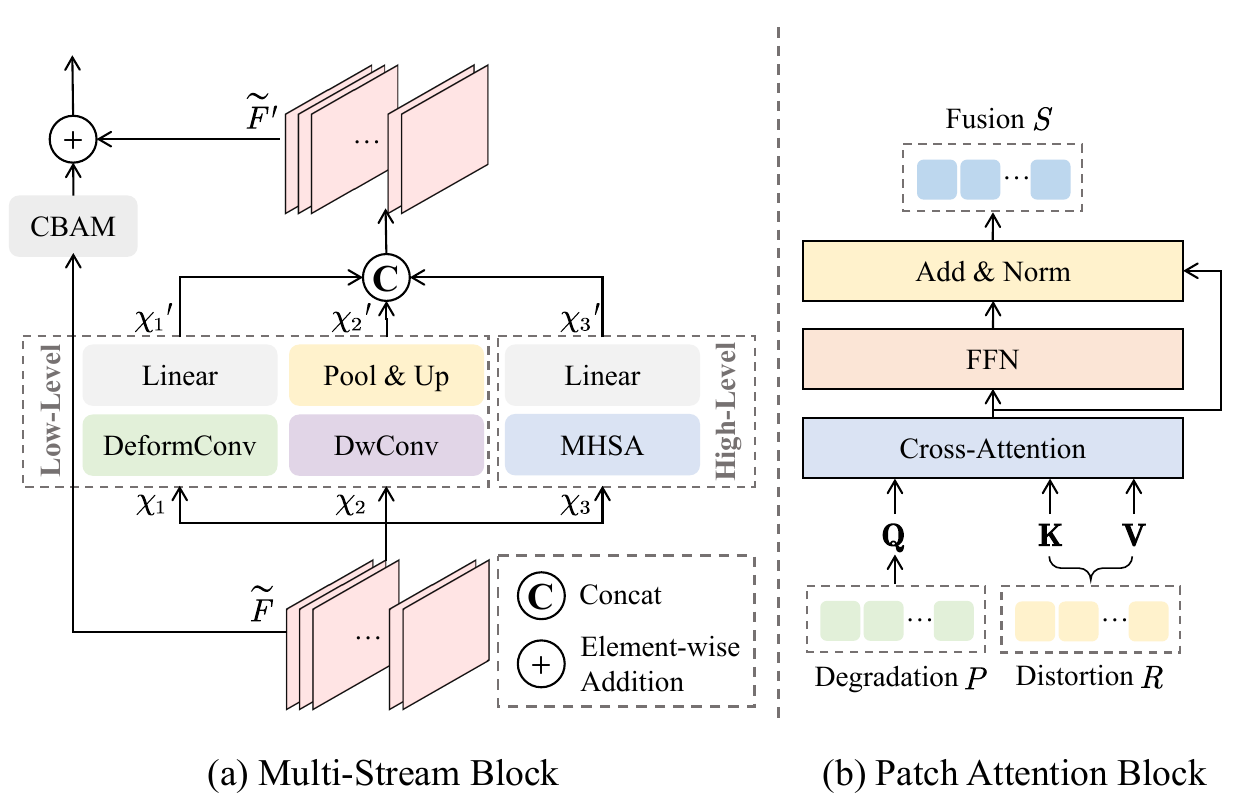}
	\caption{Overview of Multi-Stream Block and Patch Attention Block.}
	\label{fig:fig3}
 \vspace{-4mm}
\end{figure}

\subsection{Patch Attention Block\label{sec:3.4}}
According to Sec.\;\ref{sec:3.2} and Sec.\;\ref{sec:3.3}, we can obtain the distorted image features $P$ and $R$. After self-supervised pre-training on the large-scale IQA dataset KADIS, the degradation $P$ can be used to distinguish the distortion level and type of the image to be evaluated. 
As shown in Fig.\;\ref{fig:fig3} (b), 
We map $P$ to Query ($\mathbf{Q}$) and $R$ to Key ($\mathbf{K}$) and Value ($\mathbf{V}$), and then calculate them by Eq.\;(\ref{eq:eq0}) to obtain the final image quality features $S$. With this kind of computation, the model can achieve good performance in the evaluation and converge faster even on small-scale IQA datasets due to $P$-assisted training.
\begin{equation}
\label{eq:eq0}
	S = \rm {Softmax}\left(\frac{\mathbf{Q}\times \mathbf{K}^T}{\sqrt{D}}\right)\times \mathbf{V} 
\end{equation}


\subsection{Quality Prediction}
For score prediction, we first reshape the tokens from patch attention block into feature maps according to Sec.\;\ref{sec:3.4}, and then the quality score was regressed with the quality vector obtained via global average pooling (GAP). In each batch of images, we train our quality prediction model by minimizing the regression loss $\mathcal{L}$ as follows:
\begin{equation}
	\mathcal{L}=\frac{1}{N} \sum_{i=1}^{N}\left\| p_{i}-y_{i}\right\|_1
\end{equation}
where $N$ denotes the batch size, $p_{i}$ denotes the image quality score predicted by the model for the $i^{th}$ image, and $y_{i}$ denotes the corresponding objective quality score.

\section{Experiments}
\subsection{Datasets}
For self-supervised pre-training, we use the KADIS~\cite{lin2001weak} dataset. The KADIS dataset contains 140,000 original reference images, 700,000 distorted images, and has no subjective quality score. It has 25 distortion types with five distortion levels for each distortion type. In addition, we evaluated the performance of our proposed model on seven standard synthetic and authentic IQA datasets. The synthetic datasets include LIVE~\cite{sheikh2006statistical}, CSIQ~\cite{larson2010most}, TID2013~\cite{ponomarenko2015image}, and KADID-10K~\cite{lin2019kadid}, and the authentic datasets include LIVE Challenge~\cite{ghadiyaram2015massive}, KONIQ~\cite{hosu2020koniq}, and LIVE-FB~\cite{ying2020patches}. Tab.\;\ref{tab:1} shows a summary of the datasets used in the experiments. The commonly used observer ratings for the images are expressed by Mean opinion score (MOS) and Differential Mean opinion score (DMOS), where larger MOS values indicate better image quality and larger DMOS values indicate poorer image quality.

\begin{table}[h]
	\centering
	\caption{Summary of IQA datasets.}\label{tab:1}
	\label{tab:my-table}
	\scalebox{0.78}{
		\begin{tabular}{cccc}
			\toprule[1.3pt]
			Datasets & Dist. Images & Dist. types & Dataset types                     \\ \midrule
			LIVE      & 799          & 5           & Synthetic                    \\
			CSIQ      & 866          & 6           & Synthetic                    \\
			TID2013   & 3000         & 24          & Synthetic                    \\
			KADID-10K & 10125        & 25          & Synthetic                    \\
			KADIS    & 700k         & 25          & 
			Synthetic                    \\
			LIVEC     & 1162         & -           & Authentic                    \\
			KONIQ & 10073        & -           & Authentic   
            \\
            LIVE-FB   & 39810        & -           & Authentic
   \\ \bottomrule[1.3pt]
	\end{tabular}}
\end{table}

\begin{table*}[h!]
	\centering
	\caption{Comparison of SaTQA $\textit{v.s.}$ state-of-the-art NR-IQA algorithms on synthetically and authentically distorted datasets. Bold entries in \textbf{black} and \textbf{\textcolor{blue}{blue}} are the best and second-best performances, respectively. Some are borrowed from~\cite{golestaneh2022no}.}
	\label{tab:2}
	\scalebox{0.7}{
		\begin{tabular}{@{}cc|cc|cc|cc|cc|cc|cc|cc|cc@{}}
			\toprule[1.3pt]
			&              & \multicolumn{2}{c|}{LIVE}       & \multicolumn{2}{c|}{CSIQ}       & \multicolumn{2}{c|}{TID2013} & \multicolumn{2}{c|}{KADID-10K} & \multicolumn{2}{c|}{LIVE Challenge}      & \multicolumn{2}{c|}{KONIQ}      & \multicolumn{2}{c|}{LIVE-FB}     & \multicolumn{2}{c}{Average}\\ 
			&               & PLCC           & SROCC          & PLCC           & SROCC          & PLCC          & SROCC        & PLCC         & SROCC       & PLCC           & SROCC          & PLCC           & SROCC          & PLCC           & SROCC         & PLCC     & SROCC \\ \midrule
			\multicolumn{2}{c|}{DIIVINE}     & 0.908          & 0.892          & 0.776          & 0.804          & 0.567         & 0.643        & 0.435        & 0.413       & 0.591          & 0.588          & 0.558          & 0.546          & 0.187          & 0.092 & 0.575 & 0.568          \\
			\multicolumn{2}{c|}{BRISQUE}     & 0.944          & 0.929          & 0.748          & 0.812          & 0.571         & 0.626        & 0.567        & 0.528       & 0.629          & 0.629          & 0.685          & 0.681          & 0.341          & 0.303 & 0.641 & 0.644          \\
			\multicolumn{2}{c|}{ILNIQE}      & 0.906          & 0.902          & 0.865          & 0.822          & 0.648         & 0.521        & 0.558        & 0.534       & 0.508          & 0.508          & 0.537          & 0.523          & 0.332          & 0.294 & 0.622 & 0.586          \\
			\multicolumn{2}{c|}{BIECON}      & 0.961          & 0.958          & 0.823          & 0.815          & 0.762         & 0.717        & 0.648        & 0.623       & 0.613          & 0.613          & 0.654          & 0.651          & 0.428          & 0.407 & 0.698 & 0.683          \\
			\multicolumn{2}{c|}{MEON}        & 0.955          & 0.951          & 0.864          & 0.852          & 0.824         & 0.808        & 0.691        & 0.604       & 0.710          & 0.697          & 0.628          & 0.611          & 0.394          & 0.365 & 0.724 & 0.698          \\
			\multicolumn{2}{c|}{WaDIQaM}     & 0.955          & 0.960          & 0.844          & 0.852          & 0.855         & 0.835        & 0.752        & 0.739       & 0.671          & 0.682          & 0.807          & 0.804          & 0.467          & 0.455 & 0.764 & 0.761          \\
			\multicolumn{2}{c|}{DBCNN}       & \textbf{\textcolor{blue}{0.971}}          & 0.968          & 0.959          & 0.946          & 0.865         & 0.816        & 0.856        & 0.851       & 0.869          & 0.869          & 0.884          & 0.875          & 0.551          & 0.545 & 0.851 & 0.835          \\
			\multicolumn{2}{c|}{TIQA}        & 0.965          & 0.949          & 0.838          & 0.825          & 0.858         & 0.846        & 0.855        & 0.850       & 0.861          & 0.845          & 0.903          & 0.892          & 0.581          & 0.541 & 0.837 & 0.821          \\
			\multicolumn{2}{c|}{MetaIQA}     & 0.959          & 0.960          & 0.908          & 0.899          & 0.868         & 0.856        & 0.775        & 0.762       & 0.802          & 0.835          & 0.856          & 0.887          & 0.507          & 0.540 & 0.811 & 0.820          \\
			\multicolumn{2}{c|}{P2P-BM}      & 0.958          & 0.959          & 0.902          & 0.899          & 0.856         & 0.862        & 0.849        & 0.840       & 0.842          & 0.844          & 0.885          & 0.872          & 0.598          & 0.526 & 0.841 & 0.829          \\
			\multicolumn{2}{c|}{HyperIQA}    & 0.966          & 0.962          & 0.942          & 0.923          & 0.858         & 0.840        & 0.842        & 0.844       & 0.885          & \textbf{\textcolor{blue}{0.872}}          & 0.920          & 0.907          & 0.602          & 0.544 & 0.859 & 0.842          \\
			\multicolumn{2}{c|}{TReS}        & 0.968          & 0.969          & 0.942          & 0.922          & 0.883         & 0.863        & 0.858        & 0.859       & 0.877          & 0.846          & {\textbf{\textcolor{blue}{0.928}}}          & {\textbf{\textcolor{blue}{0.915}}}          & \textbf{\textcolor{blue}{0.625}}          & \textbf{\textcolor{blue}{0.554}} & 0.869 & 0.847          \\
			\multicolumn{2}{c|}{MANIQA}      & \textbf{0.983} & {\textbf{\textcolor{blue}{0.982}}}    & {\textbf{\textcolor{blue}{0.968}}}    & {\textbf{\textcolor{blue}{0.961}}}    & {\textbf{\textcolor{blue}{0.943}}}   & {\textbf{\textcolor{blue}{0.937}}}  & {\textbf{\textcolor{blue}{0.946}}}  & {\textbf{\textcolor{blue}{0.944}}} & {\textbf{\textcolor{blue}{0.887}}}        & {0.871}        & {0.915}        & {0.880}        & 0.597    & 0.543  & \textbf{\textcolor{blue}{0.891}} & \textbf{\textcolor{blue}{0.874}} \\ \midrule
			\multicolumn{2}{c|}{SaTQA (Ours)} & \textbf{0.983} & \textbf{0.983} & \textbf{0.972} & \textbf{0.965} & \textbf{0.948}    & \textbf{0.938}   & \textbf{0.949}   & \textbf{0.946}  & \textbf{0.903} & \textbf{0.877} & \textbf{0.941} & \textbf{0.930} & \textbf{0.676} & \textbf{0.582}  & \textbf{0.910} & \textbf{0.889}\\ 
			\bottomrule[1.3pt]
	\end{tabular}}
 \vspace{-3mm}
\end{table*}

\begin{table*}
	\centering
	\caption{SROCC comparisons of individual distortion types on the LIVE and CSIQ datasets.}
	\label{tab:3}
	\scalebox{0.78}{
		\begin{tabular}{c|ccccc|cccccc} 
			\toprule[1.3pt]
			Dataset  & \multicolumn{5}{c}{LIVE}              & \multicolumn{6}{c}{CSIQ}                       \\ 
			Type     & JP2K  & JPEG  & WN    & GB    & FF    & WN    & JPEG  & JP2K  & FN    & GB    & CC     \\ 
			\midrule
			BRISQUE  & 0.929 & 0.965 & 0.982 & 0.964 & 0.828 & 0.723 & 0.806 & 0.840 & 0.378 & 0.820 & 0.804  \\
			ILNIQE   & 0.894 & 0.941 & 0.981 & 0.915 & 0.833 & 0.850 & 0.899 & 0.906 & 0.874 & 0.858 & 0.501  \\
			HOSA     & 0.935 & 0.954 & 0.975 & 0.954 & 0.954 & 0.604 & 0.733 & 0.818 & 0.500 & 0.841 & 0.716  \\
			BIECON   & 0.952 & \textbf{0.974} & 0.980 & 0.956 & 0.923 & 0.902 & 0.942 & 0.954 & 0.884 & 0.946 & 0.523  \\
			WaDIQaM  & 0.942 & 0.953 & 0.982 & 0.938 & 0.923 & 0.974 & 0.853 & 0.947 & 0.882 & 0.976 & 0.923  \\
			PQR      & \textbf{0.953} & 0.965 & 0.981 & 0.944 & 0.921 & 0.915 & 0.934 & 0.955 & 0.926 & 0.921 & 0.837  \\
			HyperIQA & 0.949 & 0.961 & 0.982 & 0.926 & 0.934 & 0.927 & 0.934 & 0.960 & 0.931 & 0.915 & 0.874  \\
			MANIQA   & 0.870   & 0.895   & 0.984   & 0.959   & 0.896   & 0.966   & 0.971   & 0.973   & 0.977   & 0.956  & 0.946   \\ 
			\cmidrule{1-12}
			Ours     & 0.947 & 0.965 & \textbf{0.989} & \textbf{0.988} & \textbf{0.955} & \textbf{0.985} & \textbf{0.984} & \textbf{0.991} & \textbf{0.986} & \textbf{0.980} & \textbf{0.970}  \\
			\bottomrule[1.3pt]
	\end{tabular}}
\end{table*}

\subsection{Evaluation Metric}
We use Spearman's rank order correlation coefficient (SROCC) and Pearson's linear correlation coefficient (PLCC) to measure the performance of the NR-IQA method. SROCC is defined as follows:
\begin{equation}
	\mathrm{SROCC}=1-\frac{6 \sum_{i=1}^{n} d_{i}^{2}}{n\left(n^{2}-1\right)}
\end{equation}
where $n$ is the number of test images and $d_{i}$ denotes the difference between the ranks of $i$-th test image in ground-truth and the predicted quality scores. PLCC is defined as:
\begin{equation}
	\mathrm{PLCC}=\frac{\sum_{i=1}^{n}\left(u_{i}-\bar{u}\right)\left(v_{i}-\bar{v}\right)}{\sqrt{\sum_{i=1}^{n}\left(u_{i}-\bar{u}\right)^{2}} \sqrt{\sum_{i=1}^{n}\left(v_{i}-\bar{v}\right)^{2}}}
\end{equation}
where $u_{i}$ and $v_{i}$ denote the ground-truth and predicted quality scores of the $i$-th image, and $\bar{u}$ and $\bar{v}$ are their mean values, respectively. The values of SROCC and PLCC are in the range [-1, 1], and the higher absolute values indicate higher correlation and vice versa.

\subsection{Implementation Details}
In this paper, all of our experiments are performed using PyTorch on a single NVIDIA GeForce RTX3090. During image preprocessing, we cropped 8 random 224×224 patches from each image and randomly flipped the cropped patches horizontally. The trained patches inherit the quality score of the original images. The Encoder used for contrastive learning is a modified ResNet-50~\cite{he2016deep}, and the feature extraction part of the backbone network uses ViT-B/8~\cite{dosovitskiy2020image}, where the patch size is 8. The model is trained on ImageNet-21k and fine-tuned on ImageNet-1k. SaTQA contains three stages, the first two stages consist of Multi-Steam Block (MSB) and Patch Sampling, and the last stage contains only MSB, where $D$ is set to 768 and $h_4$ and $w_4$ are 7. In the three stages, the number of channels per branch within MSB $D_1$, $D_2$ and $D_3$ are [256, 256, 256], [192,192, 384], and [48,48,672]. The same training strategy as used in the existing NR-IQA algorithm. For the synthetic dataset, we divide the dataset based on the reference image. In the training process, we use AdamW optimizer with learning rate of $2\times 10^{-5}$, weight decay of $1\times 10^{-2}$, and learning strategy of cosine annealing, where $T_{max}$ is set to 50 and $eta_{min}$ is 0. Experiments are trained for 150 epochs. The loss function used is L1Loss, and the batch size is 4. For testing, we generate the final objective quality score by averaging the 8 cropped patch scores of the distorted images. We report the average of the results by running the experiment 10 times with different seeds.

\begin{figure*}[h]
	\centering
	\begin{minipage}{0.16\linewidth}
		\scalebox{0.15}{
			\includegraphics{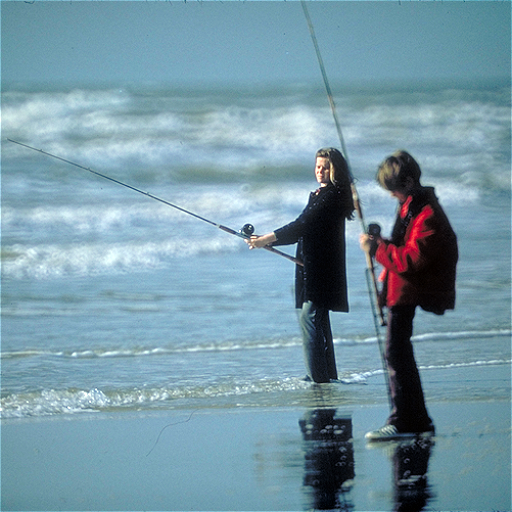}}
                \centering
                \footnotesize {DMOS $\downarrow$ \\
				MANIQA\\
				~\textbf{SaTQA(Ours)}}
		\label{fig:f-sub1}
	\end{minipage} 
	\hfill
	\begin{minipage}{0.16\linewidth}
		\scalebox{0.15}{
			\includegraphics{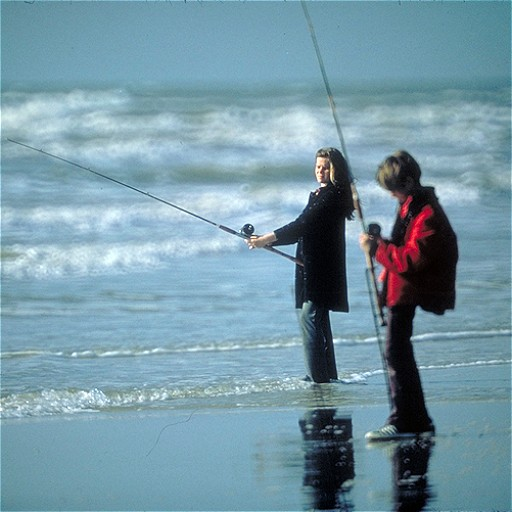}}
            \centering
		\footnotesize{0.0081~(1) \\
				0.0531~(1)\\
				~\textbf{0.0198~(1)}}
		\label{fig:f-sub2}
	\end{minipage}
	\hfill
	\begin{minipage}{0.16\linewidth}
		\scalebox{0.15}{
			\includegraphics{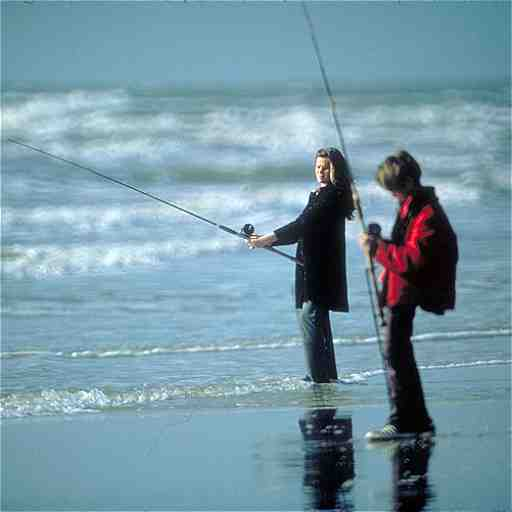}}
                \centering
		      \footnotesize{0.1805~(2) \\
				0.2074~(2)\\
				~\textbf{0.1825~(2)}}
		\label{fig:f-sub3}
	\end{minipage}
	\hfill
	\begin{minipage}{0.16\linewidth}
		\scalebox{0.15}{
			\includegraphics{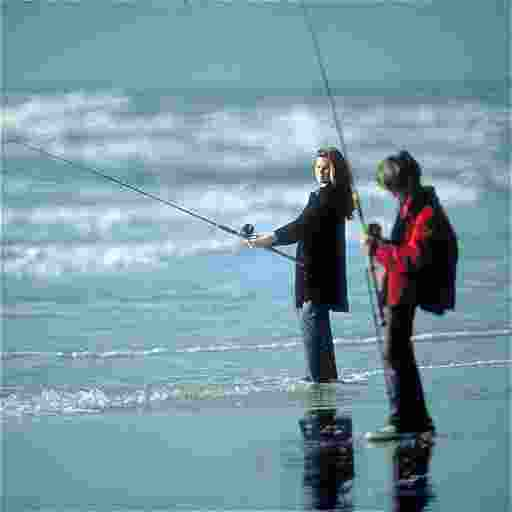}}
                \centering
		      \footnotesize{0.5178~(3) \\
				0.5179~(3)\\
				~\textbf{0.5420~(3)}}
		\label{fig:f-sub4}
	\end{minipage}
	\hfill
	\begin{minipage}{0.16\linewidth}
		\scalebox{0.15}{
			\includegraphics{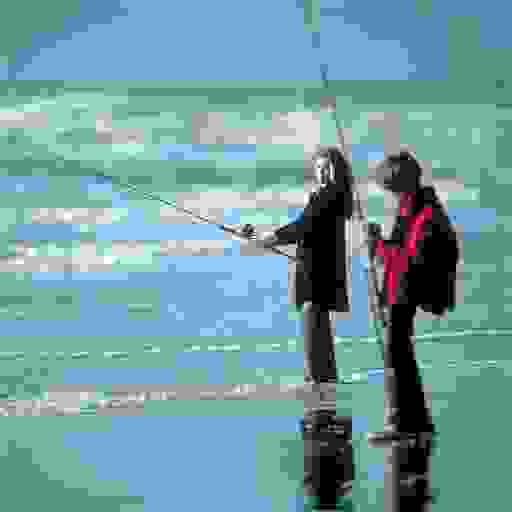}}
                \centering
		      \footnotesize{0.7246~(4) \\
				0.8139~(5)\\
				~\textbf{0.7820~(4)}}
		\label{fig:f-sub5}
	\end{minipage}	
	\hfil
	\begin{minipage}{0.17\linewidth}
		\scalebox{0.15}{
			\includegraphics{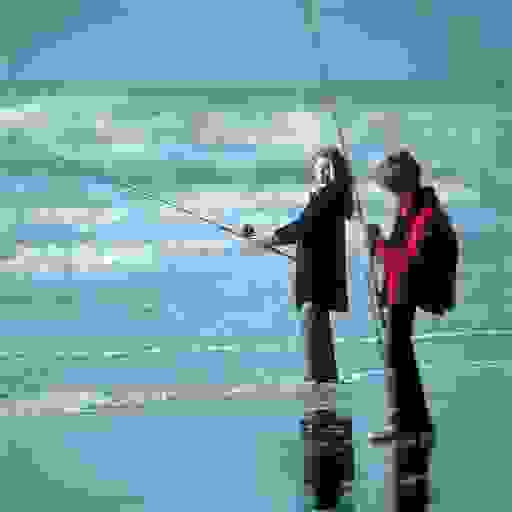}}
                \centering
        		\footnotesize{0.7907~(5) \\
				0.7549~(4) \\
				~\textbf{0.8123~(5)}}
		\label{fig:f-sub6}
	\end{minipage}	
	\vspace{+3mm}
	\caption{Example images of JPEG distortion types from the CSIQ test dataset. DMOS denotes the human rating, and smaller DMOS values indicate better image quality. The number in the parenthesis denotes the rank among considered distorted images.}
	\label{fig:fig5}
 \vspace{-3mm}
\end{figure*}

\subsection{Performance Comparison}
\noindent\textbf{Individual dataset evaluations.}
As shown in Tab.\;\ref{tab:2}, the SROCC and PLCC of SaTQA on all seven datasets are outperform of the state-of-the-art methods \cite{saad2012blind,mittal2012no,zhang2015feature,kim2016fully,ma2017end,bosse2017deep,zhang2018blind,you2021transformer,zhu2020metaiqa,ying2020patches,su2020blindly,golestaneh2022no,yang2022maniqa,xu2016blind,zeng2017probabilistic}. 
As the current correlation coefficients of the NR-IQA method on most syhthetic datasets already converge to 1 (especially the performance on LIVE approaches saturation), and it is thus extremely difficult to further make significant improvement on those datasets. Nevertheless, the performance of SaTQA on these synthetic datasets is still improved, which demonstrates the effectiveness of our proposed model. Moreover, SaTQA performs well on the authentic dataset, not only on the small-scale dataset LIVE Challenge, but also on the current large-scale LIVE-FB dataset. 

\begin{table}[h]
	\centering
	\caption{SROCC evaluations on cross datasets, where bold entries indicate the best performers.}
	\label{tab:4}
	\scalebox{0.77}{
		\begin{tabular}{c|c|c|cc} 
			\toprule[1.3pt]
			Trained on & KonIQ & LIVEC & \multicolumn{2}{c}{LIVE}  \\ 
			\cmidrule{1-5}
			Test on    & LIVEC & KONIQ & CSIQ  & TID2013           \\ 
			\midrule
			WaDIQaM    & 0.682 & 0.711 & 0.704 & 0.462             \\
			P2P-BM     & 0.770 & 0.740 & 0.712 & 0.488             \\
			HyperIQA   & 0.785 & 0.772 & 0.744 & 0.551             \\
			TReS      & 0.786 & 0.733 & 0.761 & 0.562             \\
			MANIQA     & 0.773   & 0.759   & 0.813   & 0.581               \\ 
			\cmidrule{1-5}
			Proposed   & \textbf{0.791}   & \textbf{0.788}   & \textbf{0.831}   & \textbf{0.617}               \\
			\bottomrule[1.3pt]
	\end{tabular}}
\end{table}

\noindent\textbf{Individual distortion evaluations.}
Since there are various types of distortion in images, especially in the authentic dataset, an image may contain multiple types of distortion ({e.g.}, blur, noise, {etc.}). To validate the general performance of SaTQA on the distortion types, we conducted experiments on synthetic datasets LIVE and CSIQ and compared the SROCC performance. As can be seen from Tab.\;\ref{tab:3}, our model has a strong generalization performance on CSIQ and LIVE dataset.

\noindent\textbf{Cross-dataset evaluations.}
To verify the generalization performance of the model on different datasets, we conducted cross-dataset experiments. In this validation, the training process is performed on a specific dataset and the test is executed on another different dataset without any parameter tuning during the test. As shown in Tab.\;\ref{tab:4}, our model has strong generalization ability.

\subsection{Ablation Study}
In Tab.\;\ref{tab:5}, we provide ablation studies to validate the effect of each component of our proposed model. 
We use the ViT pre-trained on ImageNet as our baseline (\#1), with the patch size of 8, and the original CLS token for quality regression.

\noindent\textbf{Supervised Contrastive Learning.}
To efficiently utilize a large amount of unlabeled data and enhance the prediction performance of the model on the authentic IQA dataset, we propose the SCL module. The results in \#2 indicate that the use of SCL can significantly improve the performance of the model on the LIVE Challenge (+0.044 PLCC, +0.037 SROCC).
For CSIQ dataset (+0.041 PLCC, +0.04 SROCC).

\noindent\textbf{Multi-Stream Block.}
To combine the feature extraction capabilities of CNN and Transformer on IQA tasks, we propose the MSB module. In \#3, the model has 0.01 improvement in SROCC and PLCC on the LIVE Challenge and CSIQ dataset.

\noindent\textbf{Patch Attention Block.}
PAB is propesed to utilize the degradation learned by SCL as a query to compute cross-attention with the perceptual distortion extracted by MSB.
From \#4, by adding PAB, the model has a 0.02 improvement in SROCC and PLCC on the LIVE Challenge dataset and a 0.002 increase on the CSIQ dataset.

\subsection{Visual Result Analysis}
We provide the DMOS prediction results of MANIQA and SaTQA for JPEG distortion images on CSIQ test set as an example. As shown in Fig.\;\ref{fig:fig5}, it can be seen that SaTQA performs better than MANIQA both in image distortion level ranking and in overall DMOS value prediction, where MANIQA produces errors in the prediction of the fourth and fifth distortion levels.

\begin{table}[]
	\centering
    \caption{Ablation study on modules. \#1 represents the baseline performance without using any proposed components.}
	\label{tab:5}
	\scalebox{0.77}{
		\begin{tabular}{c|ccc|cc|cc}
			\toprule[1.3pt]
			\multirow{2}{*}{\#} & \multirow{2}{*}{SCL}      & \multirow{2}{*}{MSB}      & \multirow{2}{*}{PAB}      & \multicolumn{2}{c|}{LIVE Challenge}                               & \multicolumn{2}{c}{CSIQ}                                          \\ 
			&                           &                           &                           & PLCC                            & SROCC                           & PLCC                            & SROCC                           \\ \midrule
			1                   &                           &                           &                           & 0.827                           & 0.803                           & 0.922                           & 0.916                           \\
			2                   & \checkmark &                           &  & 0.871                           & 0.840                           & 0.963                           & 0.956                           \\
			3                   & \checkmark & \checkmark &                           & 0.880                           & 0.851                           & 0.970                           & 0.963                           \\
		
			4                   & \checkmark & \checkmark & \checkmark & \textbf{0.903} & \textbf{0.877} & \textbf{0.972} & \textbf{0.965} \\ \bottomrule[1.3pt]
	\end{tabular}}
\end{table}


\section{Conclusion}

In this paper, we propose SaTQA, a specialized model for NR-IQA by combining supervised contrastive learning and Transformer. Initially, SCL is utilized to address incapability of existing models when facing authentic datasets for quality prediction. To enhance the feature extraction capability for distorted images and overcome the limitations of global modeling in CNN, we introduce the MSB module. Additionally, we design the PAB module to effectively fuse image degradation features and distortion information. Experimental results on seven standard IQA datasets demonstrate the remarkable performance achieved by our model.

\section*{Acknowledgements}
This work was supported in part by  the Natural Science Foundation of China (No. 62272227 and  No. 62276129), and  the Natural Science Foundation of Jiangsu Province (No. BK20220890).

\bibliography{ref}

\clearpage\textit{}


%

%

\vspace{-8mm}
\section*{Appendix}
\section{More Implementation Details}
As shown in Fig.\;\ref{fig:fig1}, for self-supervised pre-training on the KADIS~\cite{lin2001weak} dataset, the distorted and reference images are first processed through data enhancement, including random cropping, and horizontal and vertical flipping. Then the final 128-dimensional feature vectors $z_i$, $z_j$, and $z_k$ are obtained by feeding them into the modified ResNet-50~\cite{he2016deep}, respectively. Finally, the end-to-end training can be completed by applying the contrastive loss and we trained only 2 epochs. 
The temperature $\tau$ is set to 0.1, and the dimension of the final output vector in SCL is 128. 
The temperature in Eq.\,(1) regulates the sharpness of the softmax distribution, which affects the balance between positive and negative samples in contrastive learning. 
The training set includes 80\% of the data, with the remaining 20\% allocated to the test set.
Fig.\;\ref{fig:fig2} shows the detailed structure of the modified ResNet-M model.

\begin{figure}[h]
	\centering
	\includegraphics[scale=0.5]{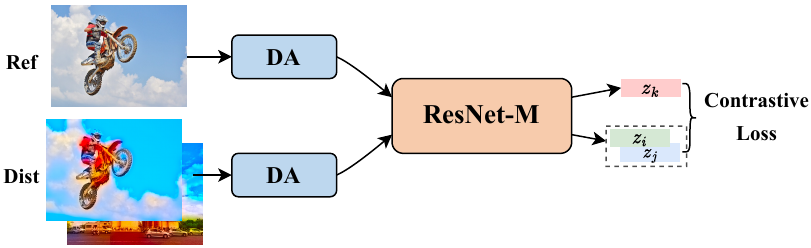}
	\caption{Self-supervised pre-training process, where \textbf{Ref} denotes the reference image, \textbf{Dist} denotes the distorted images, and \textbf{DA} denotes data augmentation.}
	 \label{fig:fig1}
	 \vspace{-4mm}
\end{figure}

\begin{figure}[h]
	\centering
	\includegraphics[scale=0.5]{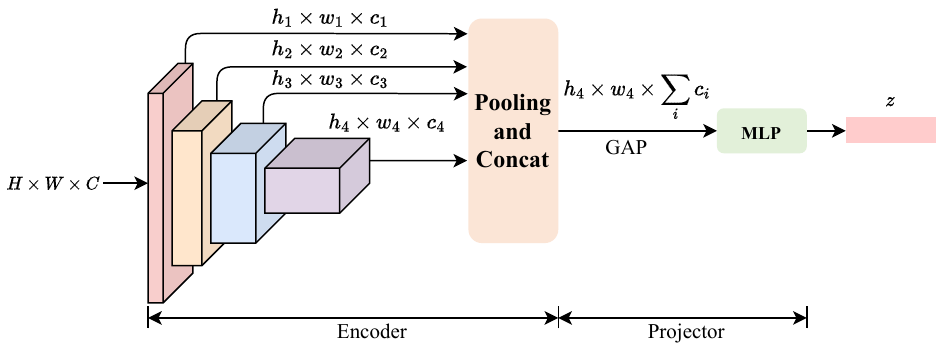}
	\caption{ResNet-M architecture.}
	 \label{fig:fig2}
	 \vspace{-4mm}
\end{figure}

\section{Model design}
\subsection{Heterogeneous structure}
To demonstrate the effectiveness of the heterogeneous model design, we implemented ablation studies on two datasets, as shown in Tab.\;\ref{tab:1}, where MSB is the proposed MSB branch in the top and MR50 indicates modified ResNet-50 in the bottom in the original paper. The results from MR50-MR50 versus MR50-MSB show a small difference in the  correlation coefficient values between using MR50 or MSB for the bottom branch, and the structure using MSB is more complex, which takes four times longer to pre-train than MR50. From the MR50-MR50 versus MSB-MR50 results, it can be seen that the use of MSB structure for the backbone network improves the model performance a lot. From the MSB-MSB versus MSB-MR50 results, it is observed that the pre-training using MR50 has better results than MSB and also lower model complexity. In summary, the best top-down choice of model structure is the combination of MSB-MR50.

\begin{table}[h]
 \vspace{-2mm}
\centering
\caption{Ablation study. MR50 indicates modified ResNet-50.} \label{tab:1}
 \vspace{-2mm}
\scalebox{0.8}{
\begin{tabular}{@{}c|cc|cc@{}}
\toprule[1.3pt]
\multirow{2}*{Top-Bottom} & \multicolumn{2}{c|}{CSIQ} & \multicolumn{2}{c}{LIVE Challenge} \\ 
                            & SROCC        & PLCC       & SROCC        & PLCC       \\ \midrule
MR50-MR50                   & 0.951        & 0.966      & 0.827        & 0.853      \\
MSB-MSB                     & 0.963        & 0.970      & 0.865        & 0.884      \\
MR50-MSB                    & 0.953        & 0.967      & 0.830        & 0.868      \\
MSB-MR50                    & 0.965        & 0.972      & 0.877        & 0.903      \\ \bottomrule[1.3pt]
\end{tabular}
}
\vspace{-4mm}
\end{table}

\subsection{Ablation on MSB structure design}
\begin{figure}[h]
	\vspace{-4mm}
	\centering
	\includegraphics[scale=0.67]{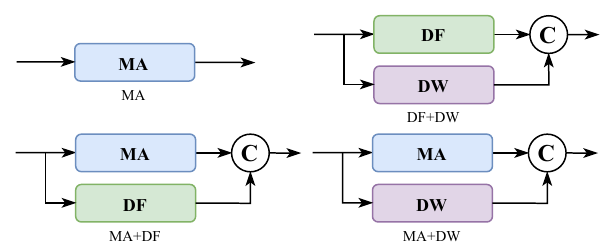}
	\caption{Different MSB structures. \textbf{``C"} means concat.}
	\label{fig:msb}
	\vspace{-4mm}
\end{figure}

To verify the efficiency of each branch of the MSB, we designed five different types of structures shown in Fig.\;\ref{fig:msb}. MA denotes the MHSA branch, DF+DW denotes the DeformConv and DwConv branches, MA+DF denotes the MHSA and DeformConv branches, MA+DW denotes the MHSA and DwConv branches, and MA+DF+DW denotes the MSB structure. Each of these structures is added with CBAM residual connection. Tab.\;\ref{tab:msb} shows the SROCC and PLCC results for the five architecture on the CSIQ and LIVE Challenge datasets.

From Tab.\;\ref{tab:msb}, it can be seen that using MA branches individually has better results than using DF+DW branches, which indicates that the attention mechanism is well adapted to the NR-IQA task. When DF or DW branches are added to the MA branch, the overall coefficient improves and the DF branch combined with MA is better than that of the DW. The overall evaluation performance of the model is the best when both DF and DW branches are added to MA.


\begin{table}[h]
\centering
\caption{Ablation study of different structures of MSB.} \label{tab:msb}
 \vspace{-2mm}
 \scalebox{0.8}{
\begin{tabular}{@{}c|cc|cc@{}}
\toprule[1.3pt]
\multirow{2}*{Architecture} & \multicolumn{2}{c|}{CSIQ} & \multicolumn{2}{c}{LIVE Challenge} \\ 
         & SROCC & PLCC & SROCC          & PLCC \\
         \midrule
MA       & 0.943   & 0.957  & 0.862            & 0.878  \\
DF+DW    & 0.936   & 0.948  & 0.846            & 0.860  \\
MA+DF    & 0.958   & 0.969  & 0.865            & 0.889  \\
MA+DW    & 0.955   & 0.963  & 0.870            & 0.886  \\
MA+DF+DW & 0.965   & 0.972  & 0.877            & 0.903  \\
\bottomrule[1.3pt]
\end{tabular}
}
\vspace{-4mm}
\end{table}

\subsection{Ablation on data amount}
\begin{table}[h!]
 \vspace{-3mm}
\centering
\caption{Performance w.r.t. different amount of data.}
\vspace{-3mm}
\label{tab:data}
\scalebox{0.8}{
\begin{tabular}{c|cc|cc}
\toprule[1.3pt]
\multirow{2}{*}{ \# of KADIS} & \multicolumn{2}{c|}{LIVE Challenge} & \multicolumn{2}{c}{CSIQ} \\
            & PLCC             & SROCC            & PLCC        & SROCC      \\ \midrule
25\%        & 0.846            & 0.838          & 0.948       & 0.937        \\
50\%        & 0.880            & 0.863          & 0.964       & 0.956        \\
100\%       & 0.903            & 0.877          & 0.972       & 0.965      \\ \bottomrule[1.3pt]
\end{tabular}
}
\vspace{-4mm}
\end{table}

Tab.\;\ref{tab:data} shows the model's performance with 50\% and 25\% of the available synthetic data, indicating that our SaTQA still achieves state-of-the-art results with only half data.

\subsection{Parameters}
Our proposed SaTQA has fewer learnable parameters (84M, 26M of which are from ResNet-50) than TReS~\cite{golestaneh2022no} (152M) and MANIQA (136M). Increasing input size may increase the number of learnable parameters, but our model still outperforms SOTA methods with a smaller number of parameters.

\section{More visualization results}

\subsection{t-SNE}
\begin{figure}[h]
	\centering
	\begin{minipage}{0.5\linewidth}
		\scalebox{0.2}{
			\includegraphics{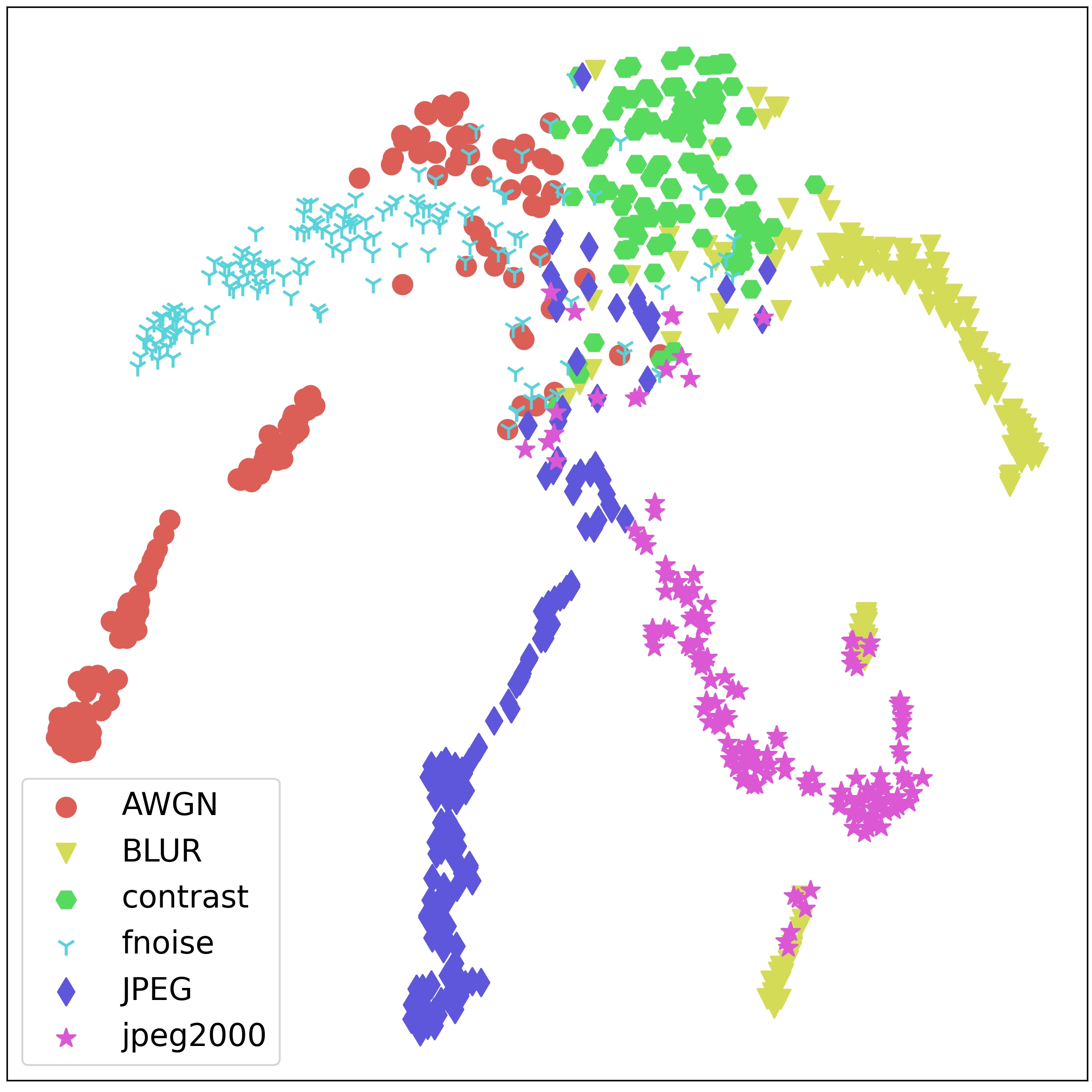}}
		\centering
            \footnotesize{(a) CSIQ w/ SCL}
		\label{fig:sub1}
	\end{minipage} 
	\hfill
	\begin{minipage}{0.49\linewidth}
		\scalebox{0.2}{
			\includegraphics{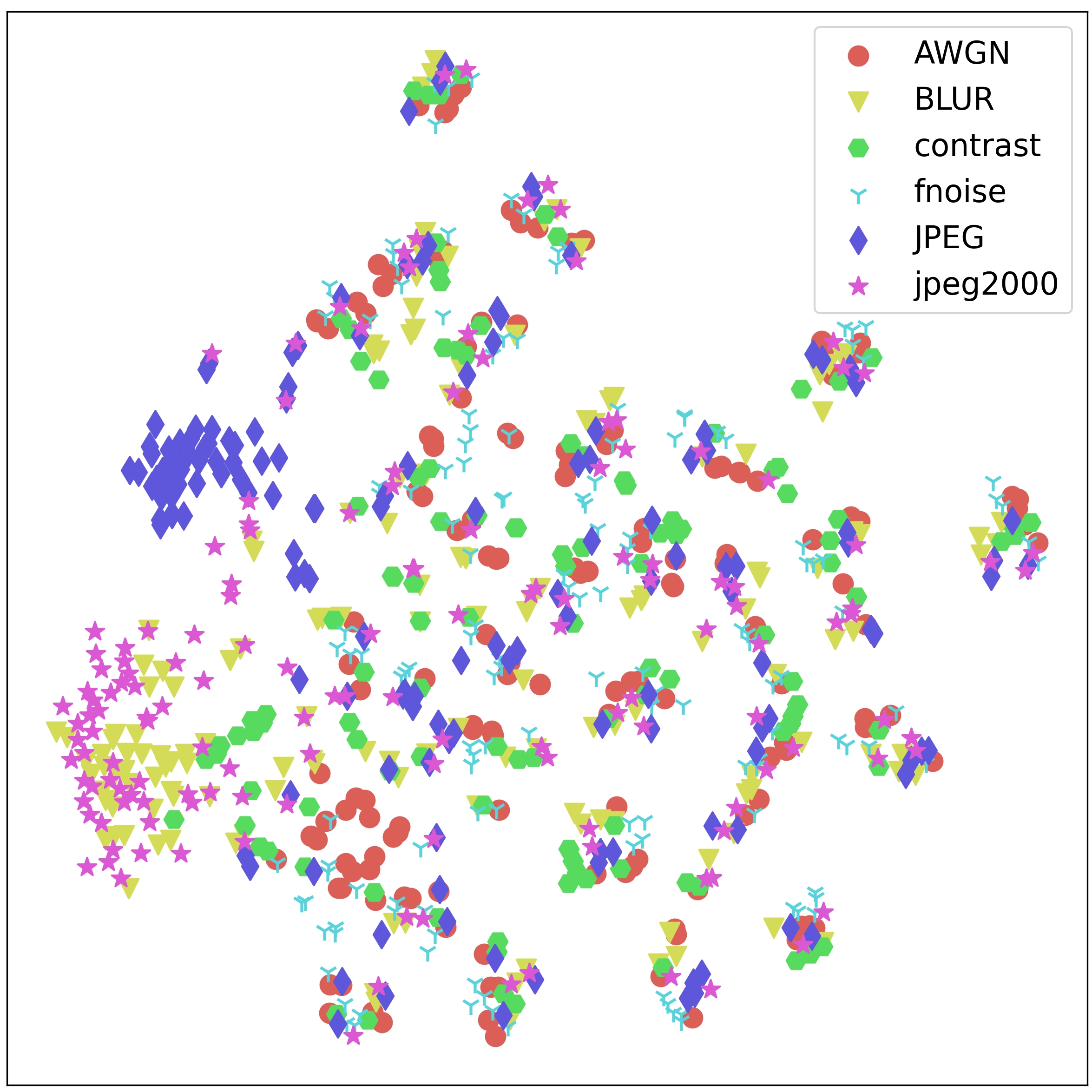}}
            \centering
		\footnotesize{(b) CSIQ w/o SCL}
		\label{fig:sub2}
	\end{minipage}
	
	\begin{minipage}{0.5\linewidth}
		\scalebox{0.2}{
			\includegraphics{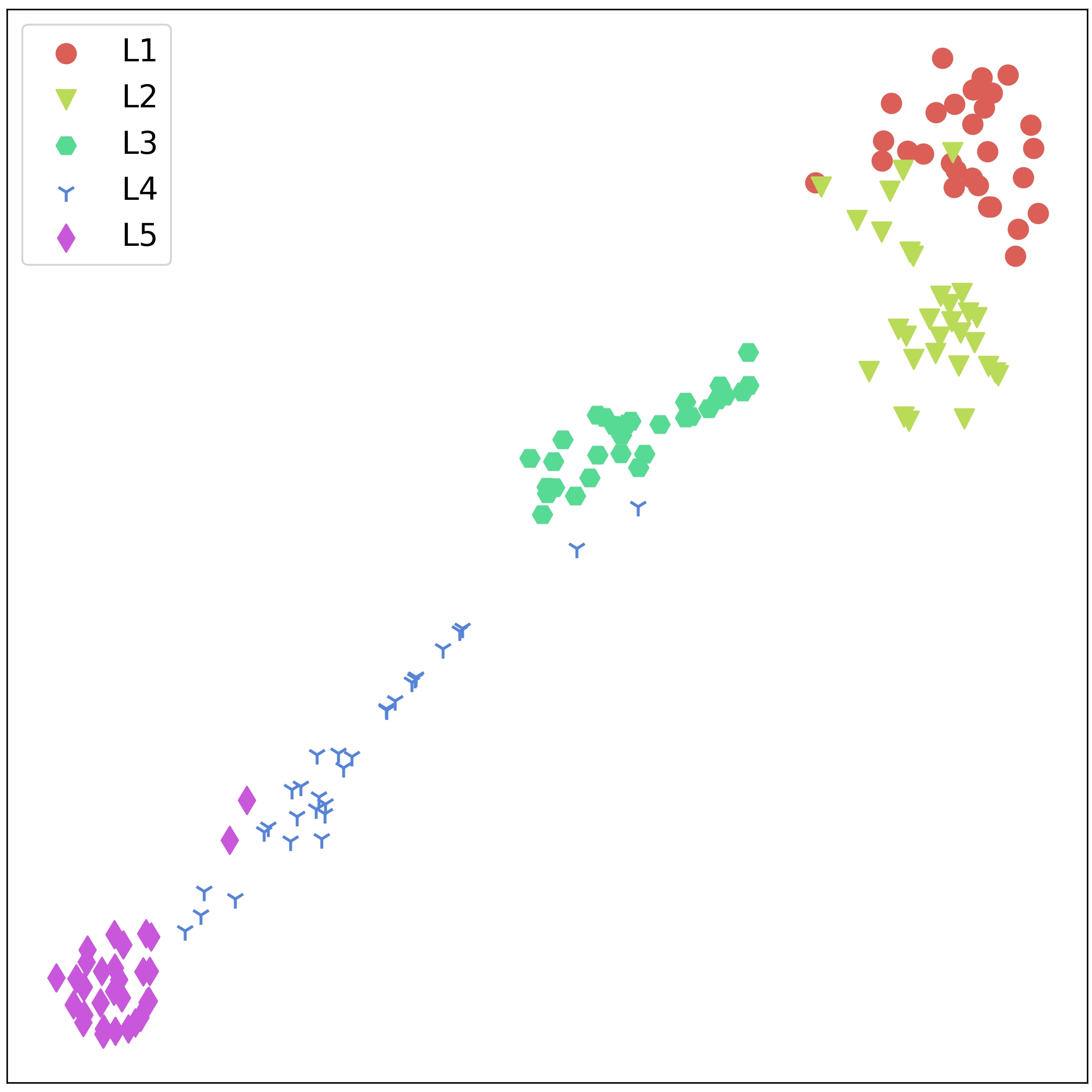}}
            \centering
		\footnotesize{(c) AWGN}
		\label{fig:sub3}
	\end{minipage}
	\hfill
	\begin{minipage}{0.49\linewidth}
		\scalebox{0.2}{
			\includegraphics{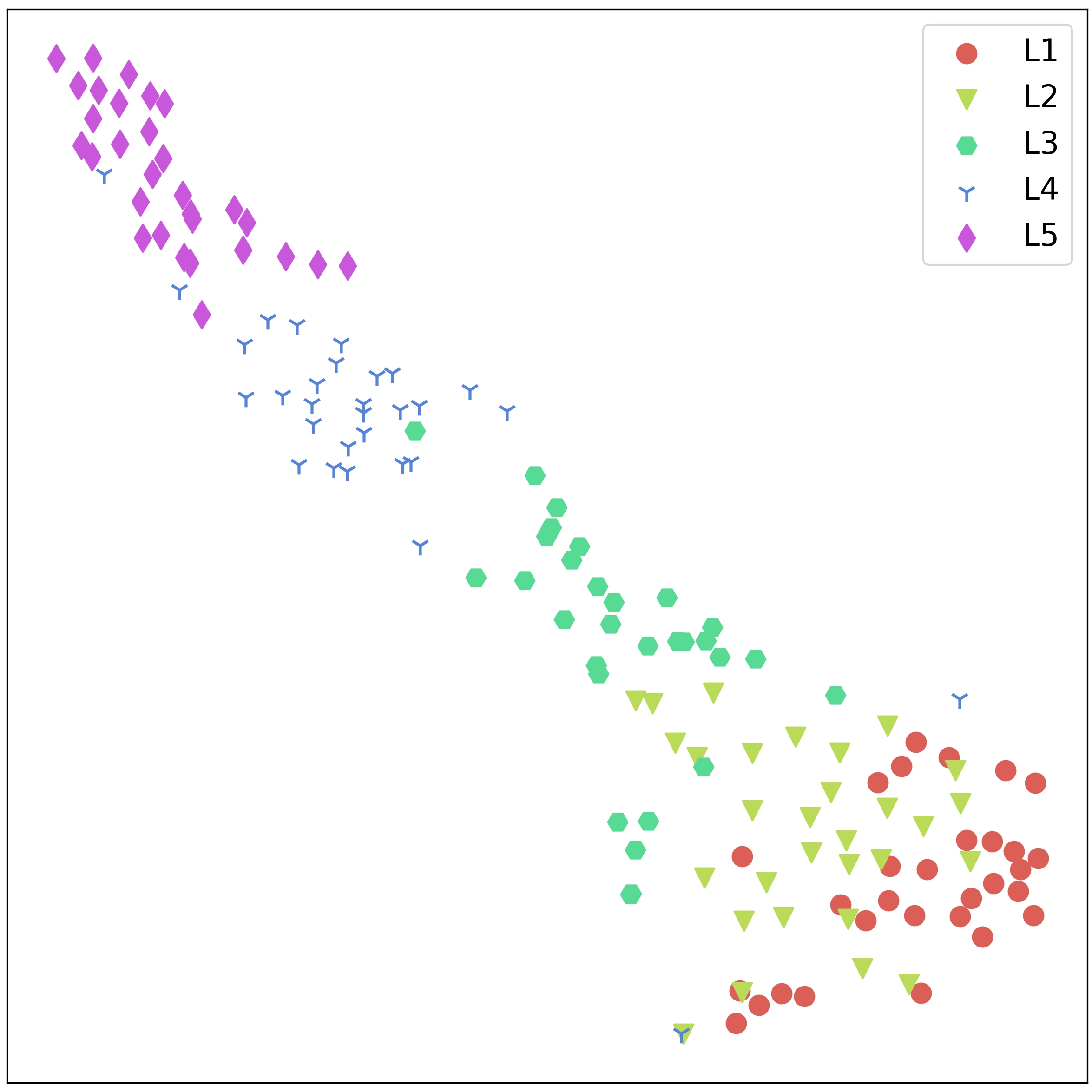}}
            \centering
		\footnotesize{(d) fnoise}
		\label{fig:sub4}
	\end{minipage}
 
	\caption{Visualization of learning degradation. (c) and (d) are the different levels of AWGN and fnoise distortion types corresponding to (a).}
	\label{fig:tsne}
\end{figure}

As shown in Fig.\;\ref{fig:tsne}, we trained the modified ResNet-50 model on the KADIS dataset using SCL, and then tested it on the CSIQ dataset with unadjusted model parameters during testing. We plotted the test results with t-SNE~\cite{van2008visualizing}, where Fig.\;10a and Fig.\;10b represent the feature clustering results of the model on the CSIQ dataset for the six distortion types. It can be seen that after self-supervised pre-training, the model is more capable of discriminating for different distortion types. Fig.\;10c and Fig.\;10d represent the visualization results of the model on five levels of specific distortion types, from which it can be seen that the model has a strong discriminating ability for each distortion level of AWGN and fnoise distortion types. This also shows the strong generalization performance of our proposed model.

\begin{figure}[h]
	\vspace{-4mm}
	\centering
	\includegraphics[scale=0.6]{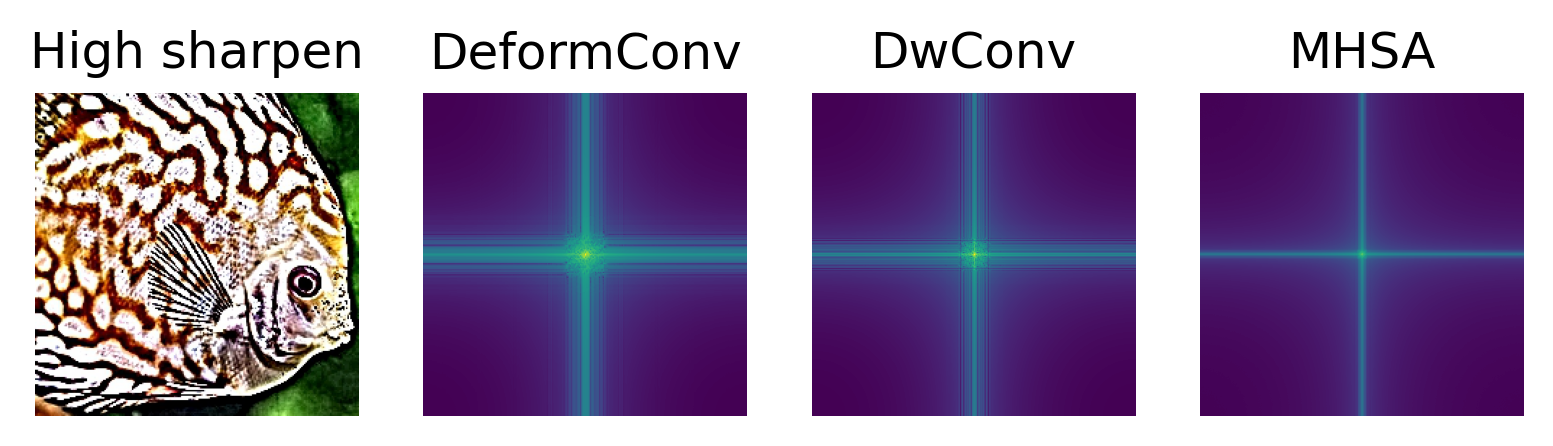}
	\caption{Fourier spectrum of DeformConv, DwConv and MHSA branches.}
	\label{fig:fourier}
	\vspace{-4mm}
\end{figure}

\begin{figure}[]
	\centering
	\begin{minipage}{0.5\linewidth}
		\scalebox{0.26}{
			\includegraphics{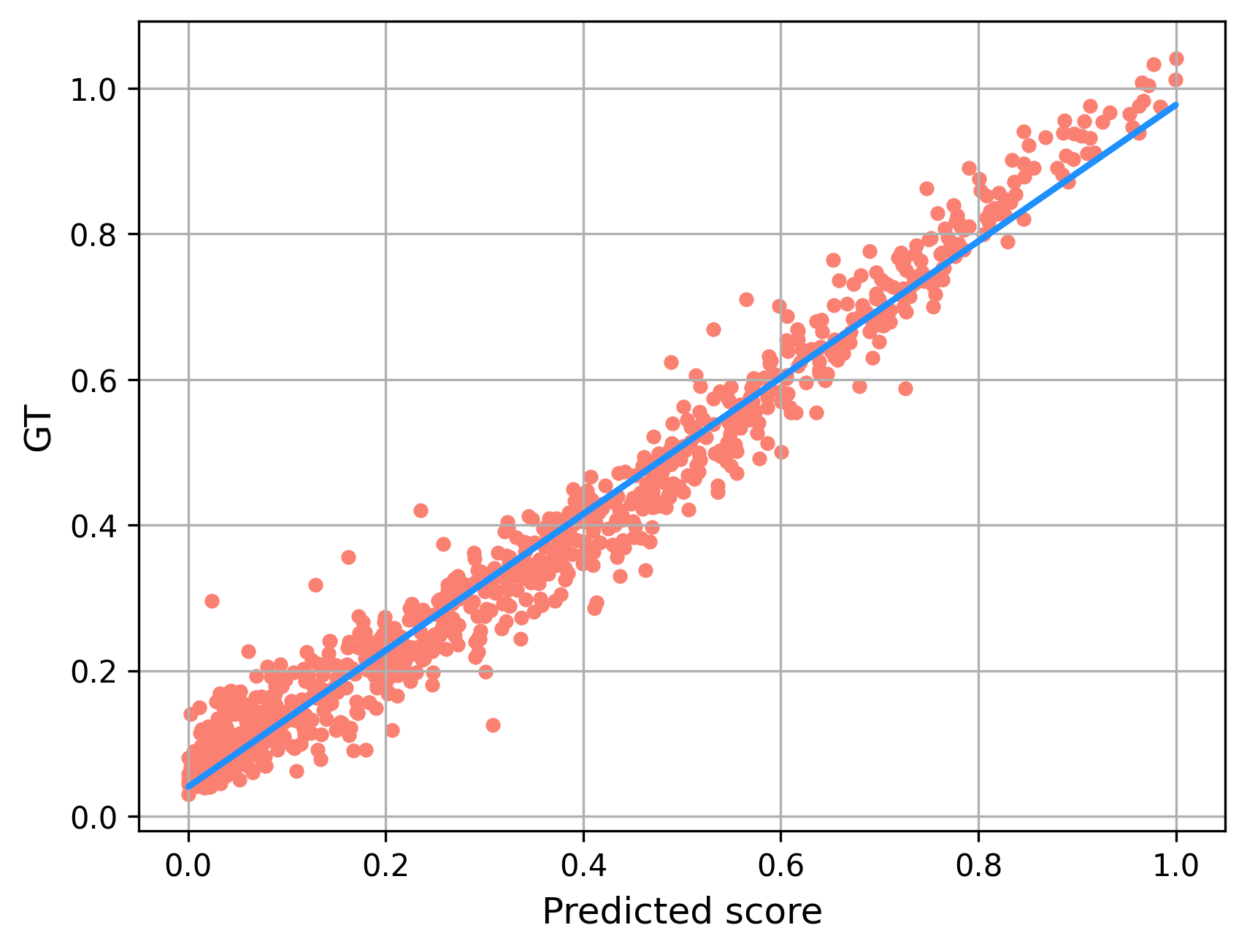}}
		\centering
		\footnotesize{(a) MANIQA on CSIQ}
		\label{fig:ff-sub1}
	\end{minipage} 
	\hfill
	\begin{minipage}{0.49\linewidth}
		\scalebox{0.26}{
			\includegraphics{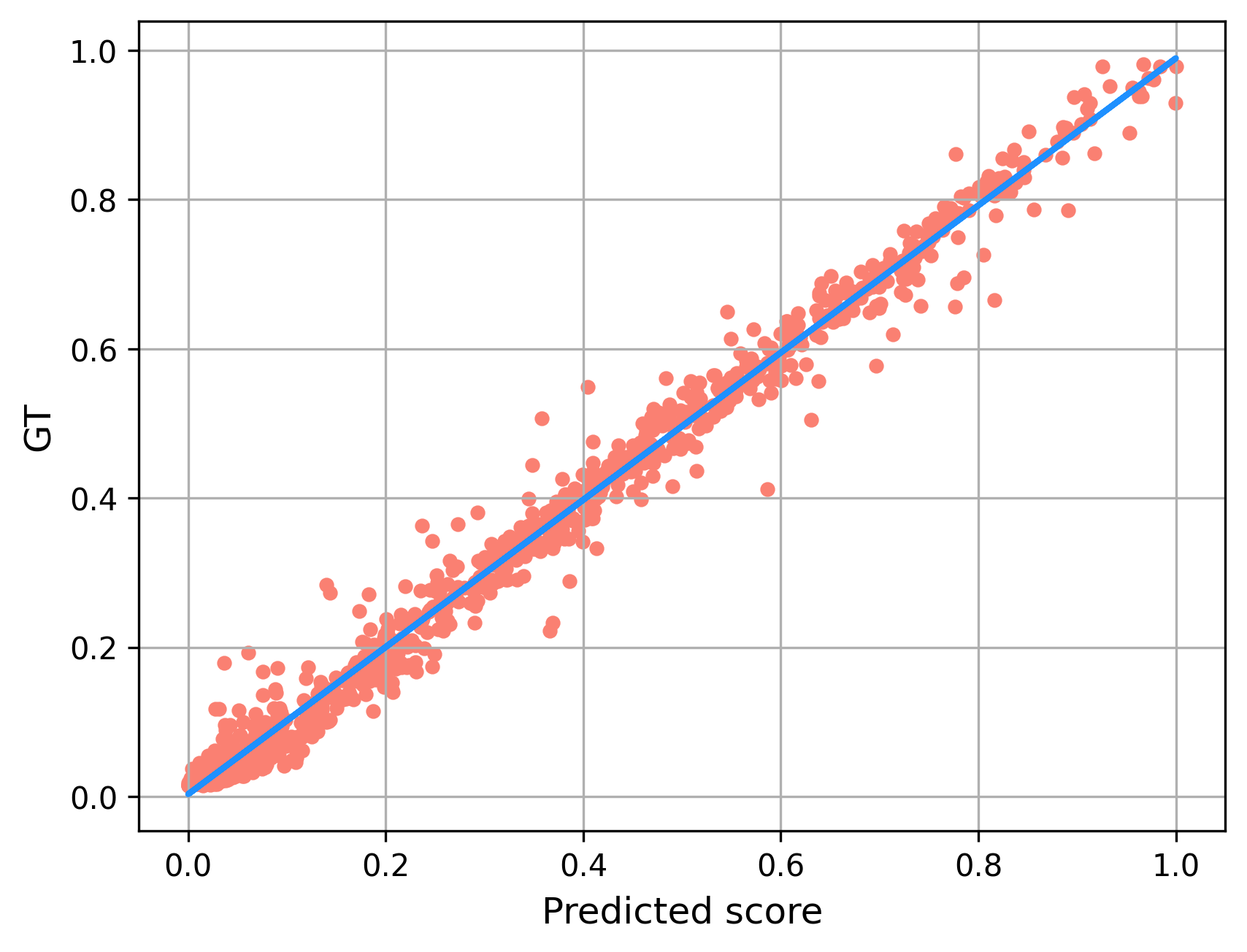}}
		\centering
		\footnotesize{(b) SaTQA on CSIQ}
		\label{fig:ff-sub2}
	\end{minipage}
	
	\begin{minipage}{0.49\linewidth}
		\scalebox{0.26}{
			\includegraphics{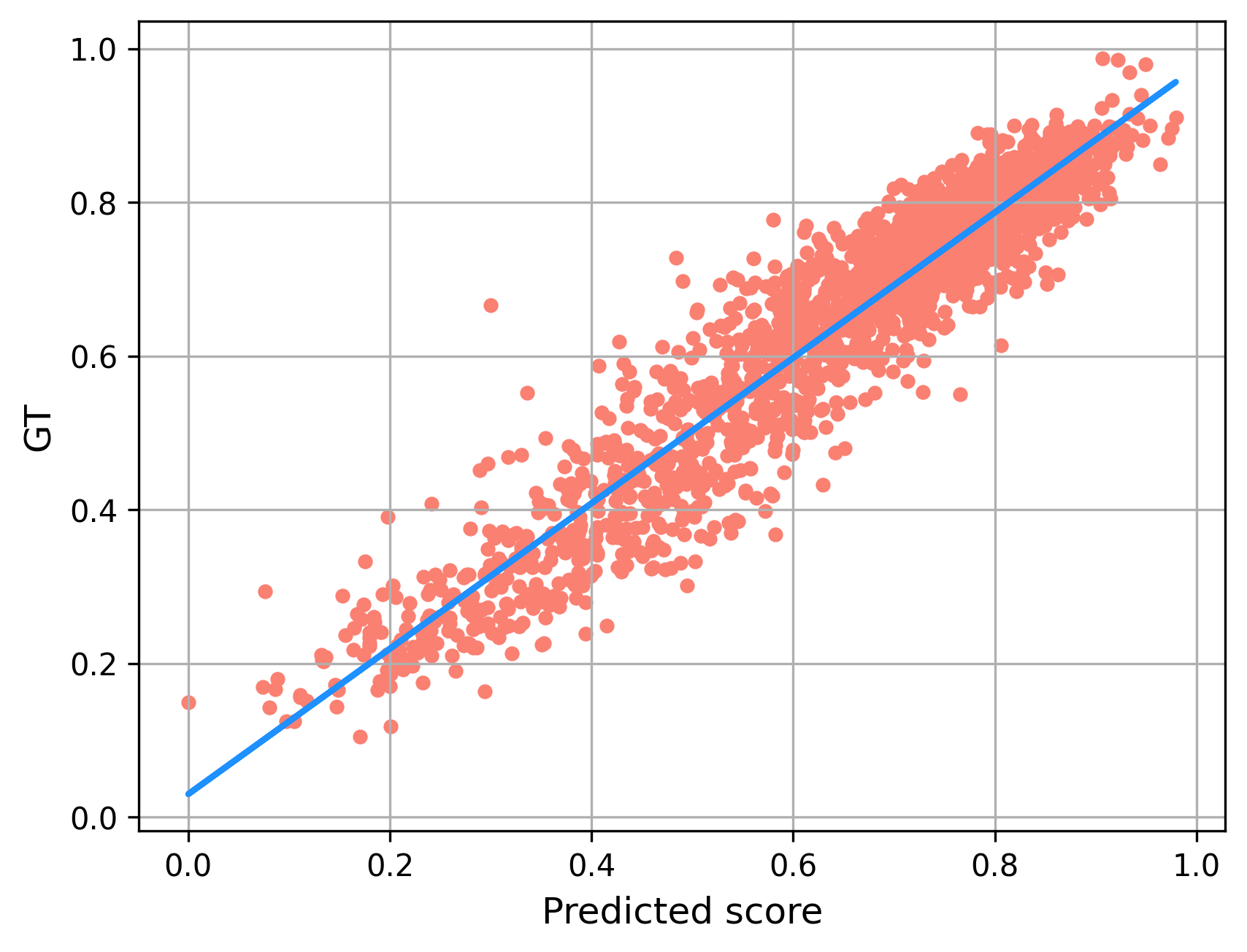}}
		\centering
		\footnotesize{(c) MANIQA on KONIQ}
		\label{fig:ff-sub3}
	\end{minipage} 
	\hfill
	\begin{minipage}{0.49\linewidth}
		\scalebox{0.26}{
			\includegraphics{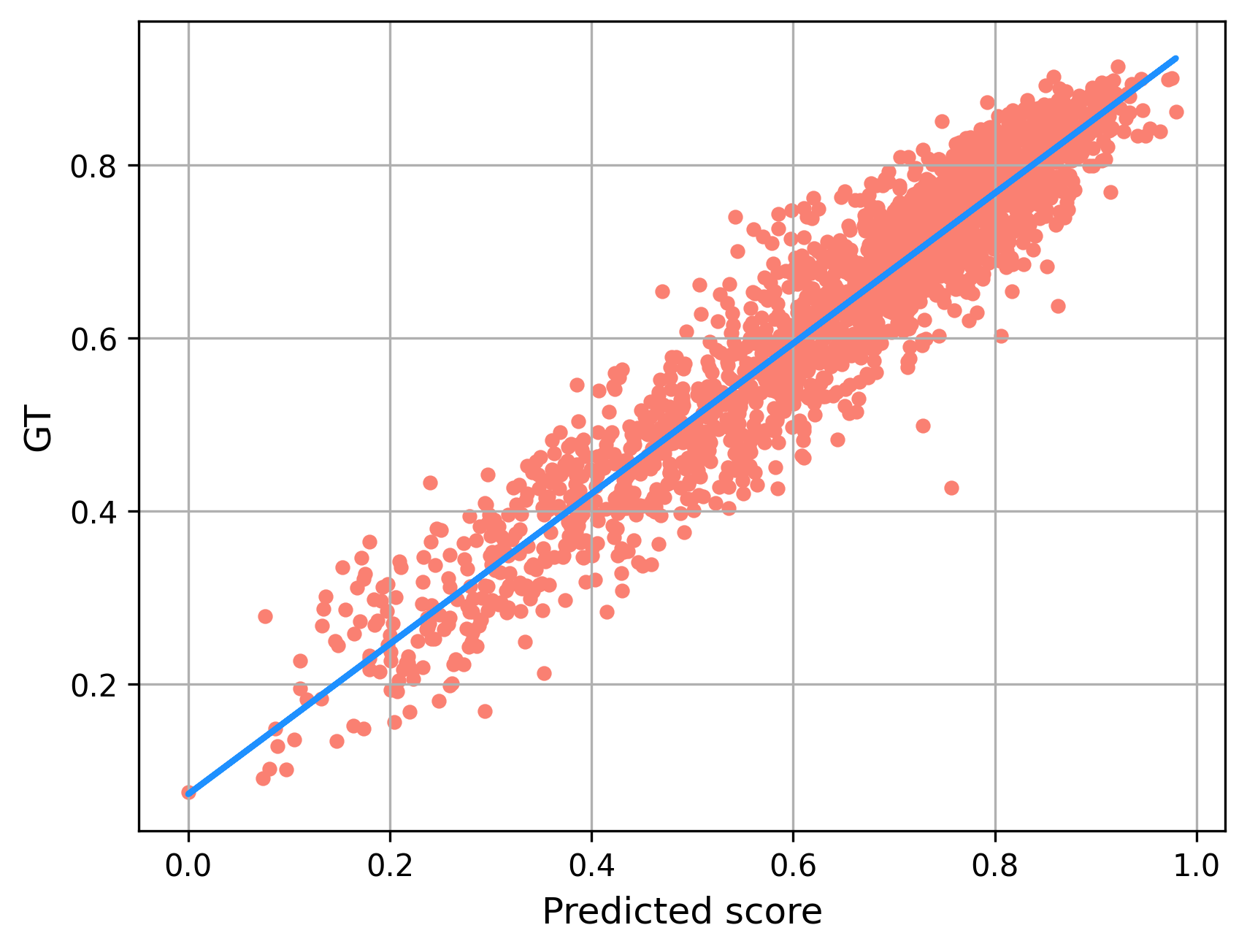}}
		\centering
		\footnotesize{(d) SaTQA on KONIQ}
		\label{fig:ff-sub4}
	\end{minipage}
	
	\caption{Scatter plots of subjective quality scores against predicted scores of MANIQA and SaTQA.}
	\label{fig:scatter}
	\vspace{-3mm}
\end{figure}

\subsection{Fourier spectrum of MSB}
In Fig.\;\ref{fig:fourier}, we visualize the fourier spectrum of the feature maps of the three branches of the MSB module. Among them, the MHSA branch focuses more on the low frequency region, while the DeformConv and DwConv branches focus mainly on the high frequency region, which demonstrates that CNN combined with attention can extract more high and low frequency features on images.

\subsection{Scatter plots}

In Fig.\;\ref{fig:scatter}, we scatter plotted and linearly fitted the prediction scores and ground-truth scores of MANIQA \cite{yang2022maniqa} and SaTQA on the CSIQ and KONIQ~\cite{hosu2020koniq} datasets. On the CSIQ dataset, the prediction results of our proposed model are more linearly changing with the true values. On the KONIQ dataset, MANIQA would have more outliers, while our model results are more concentrated.

\subsection{More comparison of DMOS value prediction results}
Fig. \ref{fig:fig3} to \ref{fig:fig4} show more prediction results of DMOS values for MANIQA and SaTQA on the CSIQ dataset, and it can be seen that the evaluation performance of our model is obviously better.

\begin{figure*}[]
	\centering
	\captionsetup[subfigure]{labelformat=empty}
	\begin{subfigure}{0.16\linewidth}
		\scalebox{0.15}{
			\includegraphics{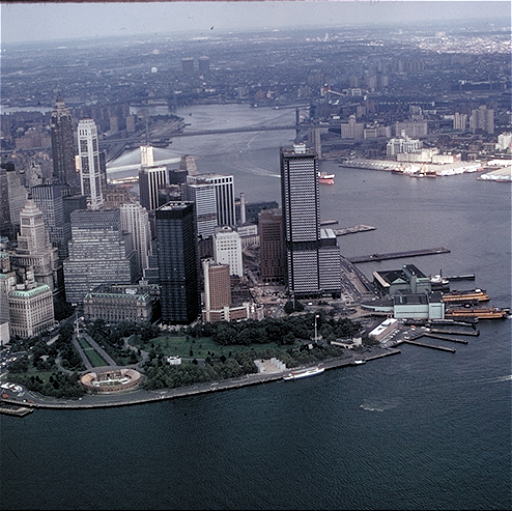}}
		\captionsetup{justification=centering}
		\caption{\textbf{DMOS $\downarrow$ \\
				MANIQA\\
				SaTQA(Ours)}}
	\end{subfigure} 
	\hfill
	\begin{subfigure}{0.16\linewidth}
		\scalebox{0.15}{
			\includegraphics{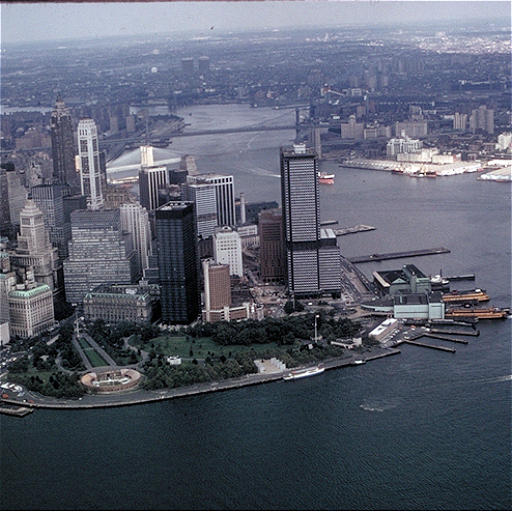}}
		\captionsetup{justification=centering}
		\caption{\textbf{0.0316~(1) \\
				0.1092~(1)\\
				0.0283~(1)}}
	\end{subfigure}
	\hfill
	\begin{subfigure}{0.16\linewidth}
		\scalebox{0.15}{
			\includegraphics{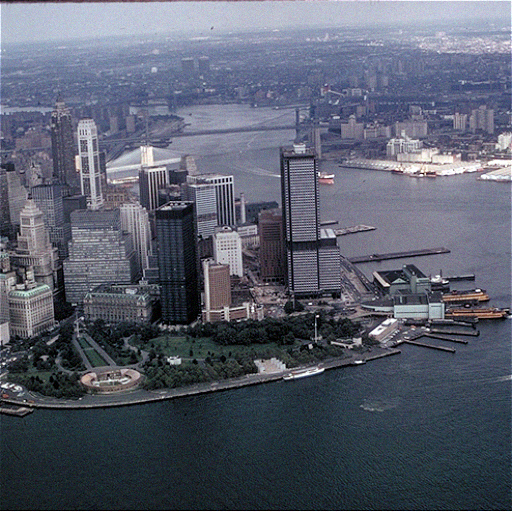}}
		\captionsetup{justification=centering}
		\caption{\textbf{0.1251~(2) \\
				0.1747~(2)\\
				0.1339~(2)}}
	\end{subfigure}
	\hfill
	\begin{subfigure}{0.16\linewidth}
		\scalebox{0.15}{
			\includegraphics{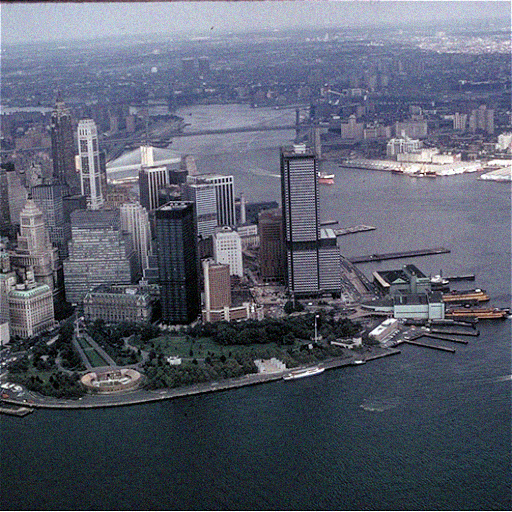}}
		\captionsetup{justification=centering}
		\caption{\textbf{0.2450~(3) \\
				0.2262~(3)\\
				0.2623~(3)}}
	\end{subfigure}
	\hfill
	\begin{subfigure}{0.16\linewidth}
		\scalebox{0.15}{
			\includegraphics{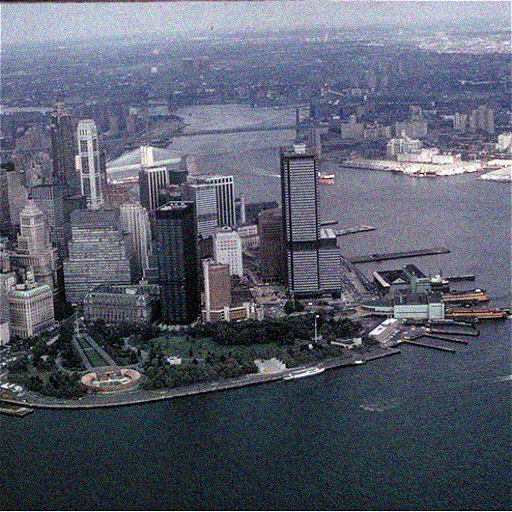}}
		\captionsetup{justification=centering}
		\caption{\textbf{0.3730~(4) \\
				0.3671~(4)\\
				0.3707~(4)}}
	\end{subfigure}	
	\hfil
	\begin{subfigure}{0.17\linewidth}
		\scalebox{0.15}{
			\includegraphics{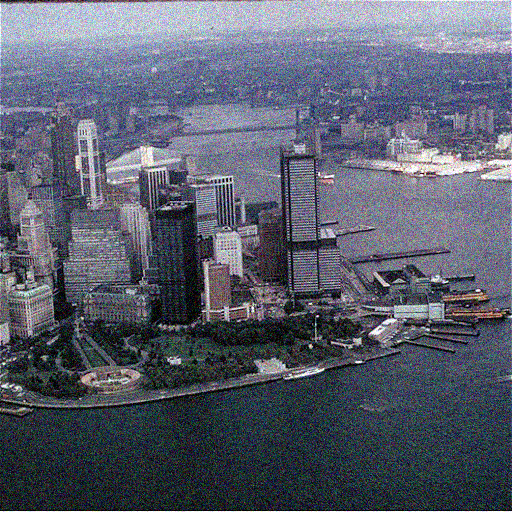}}
		\captionsetup{justification=centering}
		\caption{\textbf{0.5012~(5) \\
				0.4468~(5)\\
				0.5091~(5)}}
	\end{subfigure}	
	
	\caption{Example images of AWGN distortion types from the CSIQ test dataset. The number in the parenthesis denotes the rank among considered distorted images in this figure.}
	\label{fig:fig3}
\end{figure*}

\begin{figure*}[]
	\centering
	\captionsetup[subfigure]{labelformat=empty}
	\begin{subfigure}{0.16\linewidth}
		\scalebox{0.15}{
			\includegraphics{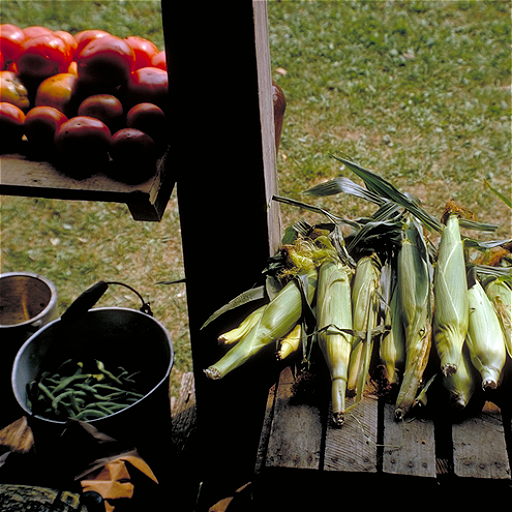}}
		\captionsetup{justification=centering}
		\caption{\textbf{DMOS $\downarrow$ \\
				MANIQA\\
				SaTQA(Ours)}}
	\end{subfigure} 
	\hfill
	\begin{subfigure}{0.16\linewidth}
		\scalebox{0.15}{
			\includegraphics{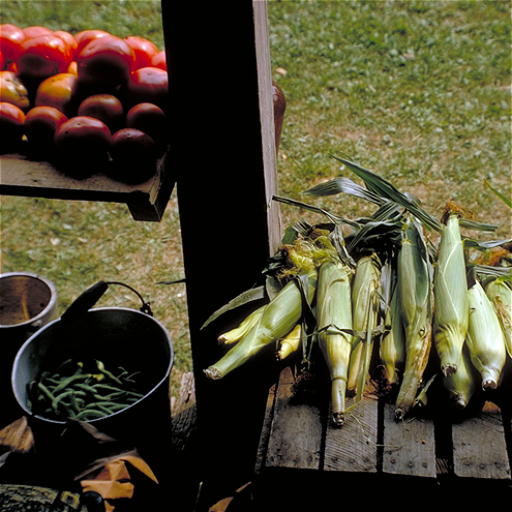}}
		\captionsetup{justification=centering}
		\caption{\textbf{0.0564~(1) \\
				0.1003~(1)\\
				0.0570~(1)}}
	\end{subfigure}
	\hfill
	\begin{subfigure}{0.16\linewidth}
		\scalebox{0.15}{
			\includegraphics{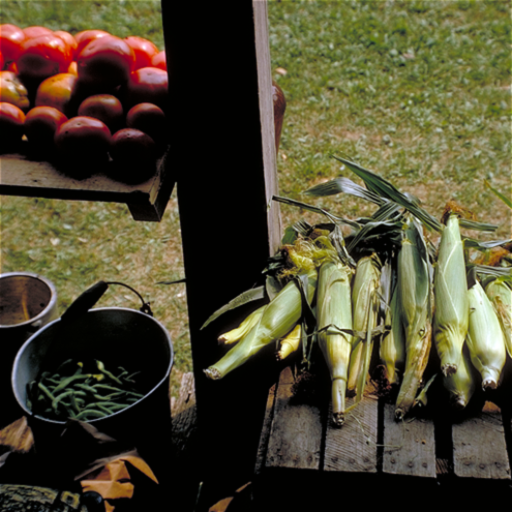}}
		\captionsetup{justification=centering}
		\caption{\textbf{0.2063~(2) \\
				0.1161~(2)\\
				0.1497~(2)}}
	\end{subfigure}
	\hfill
	\begin{subfigure}{0.16\linewidth}
		\scalebox{0.15}{
			\includegraphics{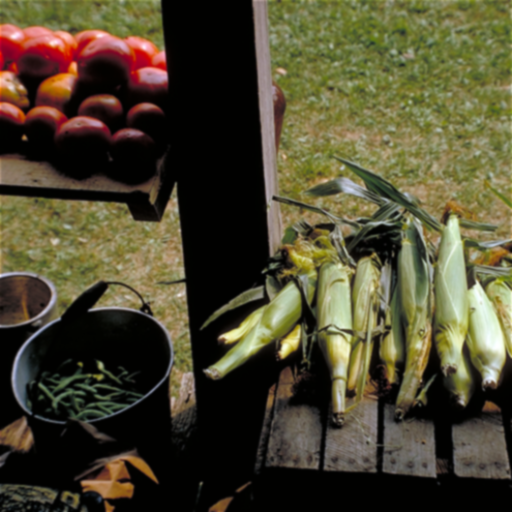}}
		\captionsetup{justification=centering}
		\caption{\textbf{0.3489~(3) \\
				0.3394~(3)\\
				0.3548~(3)}}
	\end{subfigure}
	\hfill
	\begin{subfigure}{0.16\linewidth}
		\scalebox{0.15}{
			\includegraphics{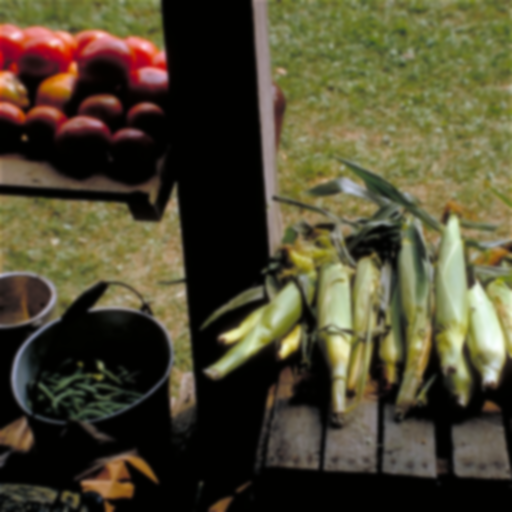}}
		\captionsetup{justification=centering}
		\caption{\textbf{0.5214~(4) \\
				0.5543~(4)\\
				0.5275~(4)}}
	\end{subfigure}	
	\hfil
	\begin{subfigure}{0.17\linewidth}
		\scalebox{0.15}{
			\includegraphics{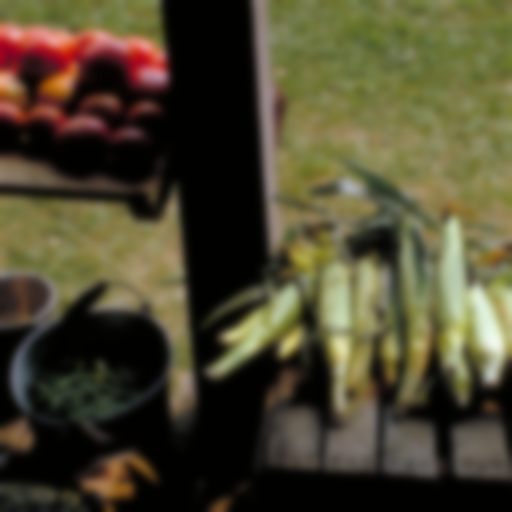}}
		\captionsetup{justification=centering}
		\caption{\textbf{0.7188~(5) \\
				0.7716~(5)\\
				0.7077~(5)}}
	\end{subfigure}	
	
	\caption{Example images of BLUR distortion types from the CSIQ test dataset. The number in the parenthesis denotes the rank among considered distorted images in this figure.}
	\label{fig:fig4}
\end{figure*}

\subsection{Grad-CAM}
Fig. \ref{fig:fig5} show the Grad-CAM~\cite{selvaraju2017grad} activation maps of SaTQA on the TID2103 dataset. The distorted images are shown at the top and the corresponding cam maps are shown at the bottom, where the brighter areas indicate that the model is more concerned. It can be seen that the perception of the distortion region by our proposed model is more consistent with the visual sensation of human eyes.

\begin{figure*}[]
	\centering
	\captionsetup[subfigure]{labelformat=empty}
	\begin{subfigure}{0.19\linewidth}
		\scalebox{0.75}{
			\includegraphics{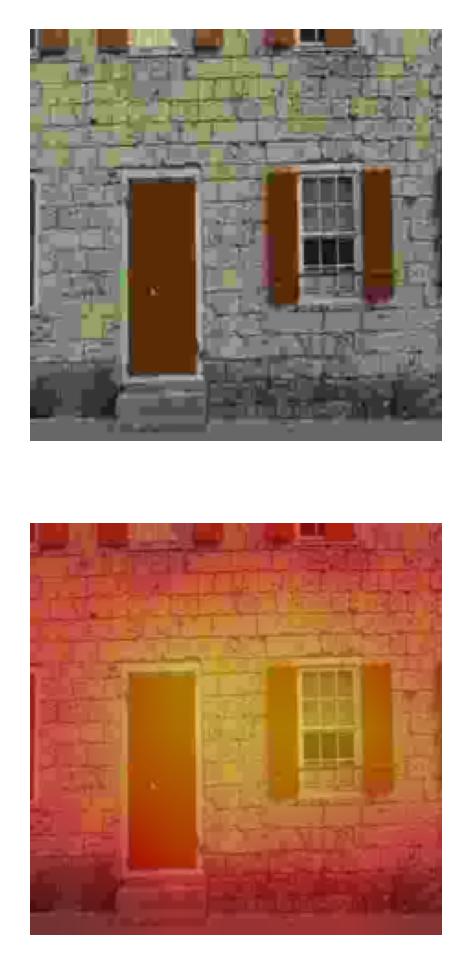}}
		\end{subfigure} 
		\hfill
		\begin{subfigure}{0.19\linewidth}
			\scalebox{0.75}{
				\includegraphics{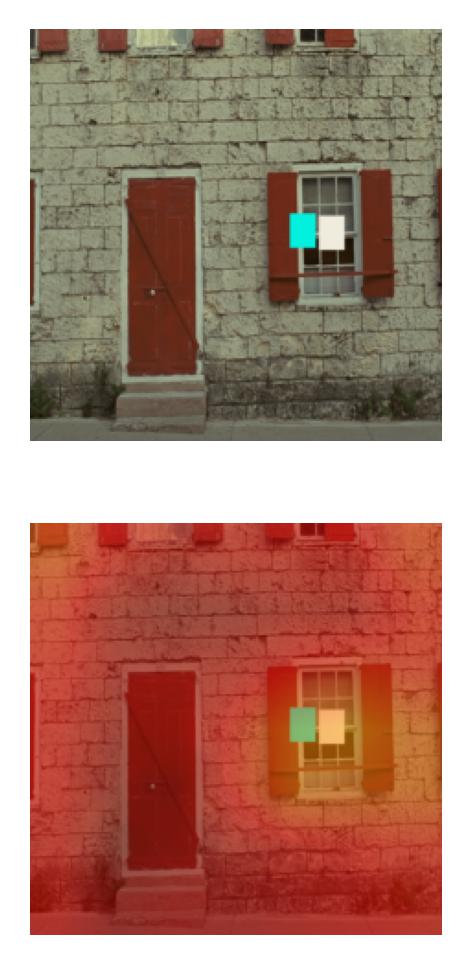}}
		\end{subfigure}
		\hfill
		\begin{subfigure}{0.19\linewidth}
			\scalebox{0.75}{
				\includegraphics{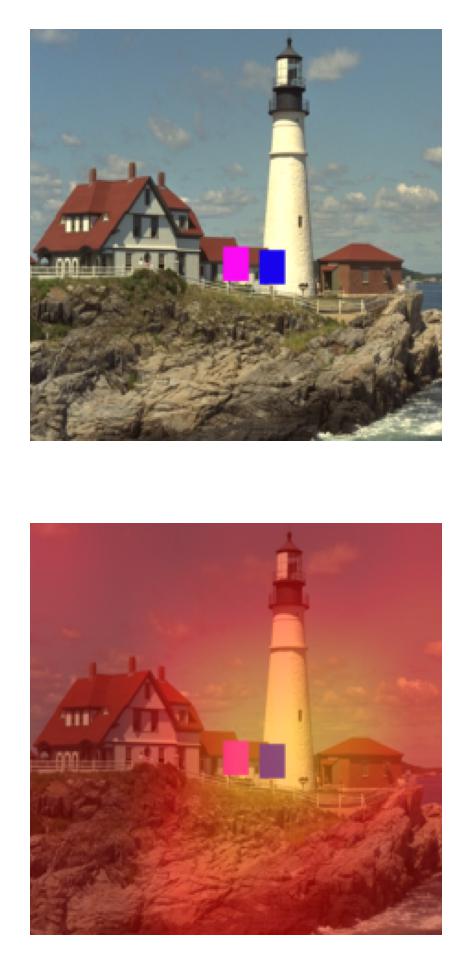}}
		\end{subfigure}
		\hfill
		\begin{subfigure}{0.19\linewidth}
			\scalebox{0.75}{
				\includegraphics{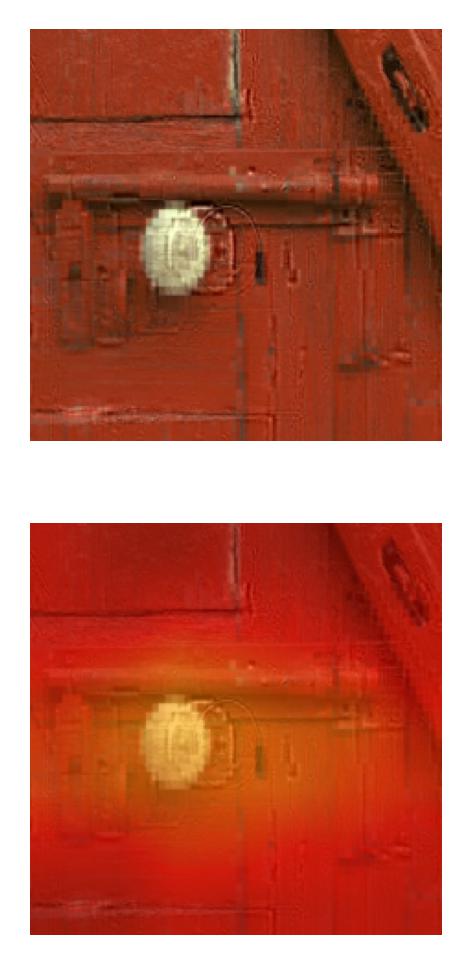}}
		\end{subfigure}
		\hfill
		\begin{subfigure}{0.21\linewidth}
			\scalebox{0.75}{
				\includegraphics{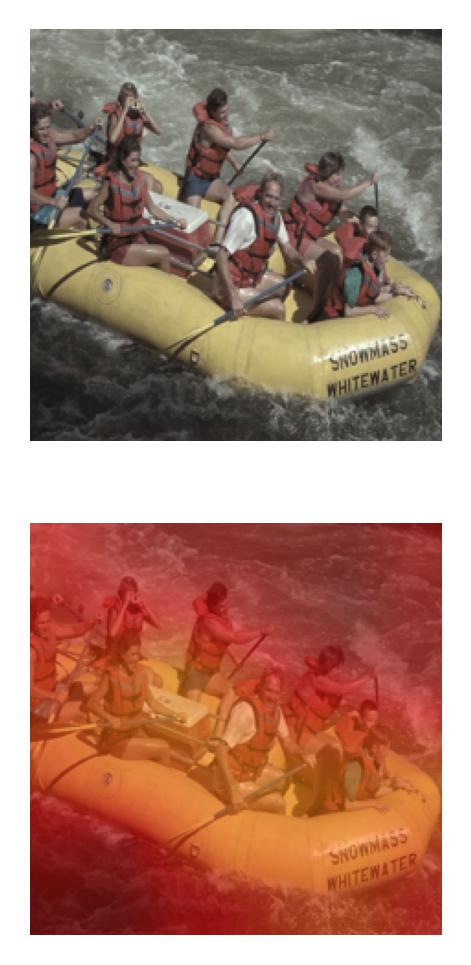}}
		\end{subfigure}	
		\hfill
		
		\caption{Grad-CAM activation maps of SaTQA on TID2013 dataset.}
		\label{fig:fig5}
	\end{figure*}

\end{document}